\documentclass[10pt,journal,compsoc]{IEEEtran}
\ifCLASSOPTIONcompsoc
  \usepackage[nocompress]{cite}
\else
  \usepackage{cite}
\fi
\usepackage{graphicx}
\usepackage{amsmath}
\usepackage{amssymb, url}
\usepackage{bm}
\usepackage[usenames]{color}
\usepackage{icomma}
\usepackage[ruled,lined]{algorithm2e}
\usepackage[svgnames,table]{xcolor}
\usepackage{makecell}
\usepackage{booktabs}
\usepackage{multirow}
\usepackage[bf,font=footnotesize]{caption}
\usepackage{array}
\usepackage{wrapfig}

\newcommand{\etal}{\emph{et al.}}

\newcommand{\ie}{\emph{i.e.}}

\def\Att{{V_{att}}}
\def\Cap{{V_{cap}}}
\def\Know{{V_{know}}}
\def\V2L{\textit{V2L}}
\def\CNN{\mbox{CNN}}
\def\ModParms{{\boldsymbol \theta}}

\begin{document}
\title{Image Captioning and Visual Question Answering Based on Attributes
\\and External Knowledge}
\author{Qi Wu, Chunhua Shen,  Peng Wang, Anthony Dick, Anton van den Hengel
\IEEEcompsocitemizethanks{\IEEEcompsocthanksitem
  The authors are with the Australian Centre for Visual Technologies, and  School of Computer Science,
  at The University of Adelaide, Australia.
  E-mail: (\{qi.wu01, chunhua.shen, p.wang, anthony.dick, anton.vandenhengel\}@\-adelaide.\-edu.\-au).
}
}

\markboth{Manuscript, 2016}%
{}

\IEEEtitleabstractindextext{%

\begin{abstract}
Much of the recent progress in Vision-to-Language problems has been achieved through a combination of Convolutional Neural Networks (CNNs)
and Recurrent Neural Networks (RNNs). This approach does not explicitly represent high-level semantic concepts, but rather seeks to progress directly from image features to text. In this paper we first propose a method of incorporating high-level concepts into the successful CNN-RNN approach, and show that it achieves a significant improvement on the state-of-the-art in both image captioning and visual question answering.
We further show that the same mechanism can be used to incorporate external knowledge,
which is critically important for answering high level visual questions.
Specifically, we design a visual question answering model that combines an internal representation of the content of an
image with information extracted from a general knowledge base to answer a broad range of image-based questions.
It particularly allows questions to be asked 
where the image alone does not contain the the information required to select the appropriate answer.
Our final model achieves the best reported results for both image captioning and visual question answering on several of the major benchmark datasets.
\end{abstract}

}
\maketitle

\IEEEdisplaynontitleabstractindextext
\IEEEpeerreviewmaketitle

\section{Introduction}
\label{sec:introduction}
\IEEEPARstart{V}ision-to-Language problems present a particular challenge in Computer Vision because they require translation between two different forms of information.  In this sense the problem is similar to that of machine translation between languages. In machine language translation there have been a series of results showing that good performance can be achieved without developing a higher-level model of the state of the world. In~\cite{bahdanau2014neural,cho2014learning,sutskever2014sequence}, for instance, a source sentence is transformed into a fixed-length vector representation by an `encoder' RNN, which in turn is used as the initial hidden state of a `decoder' RNN that generates the target sentence.

Despite the supposed equivalence between an image and  a thousand words, the
manner in which information is represented in each data form could hardly be
more different. Human language is designed specifically so as to communicate
information between humans, whereas even the most carefully composed image is
the culmination of a complex set of physical processes over which humans have
little control.  Given the differences between these two forms of information,
it seems surprising that methods inspired by machine language translation have
been so successful. These RNN-based methods which translate directly from image
features to text, without developing a high-level model of the state of the
world, represent the current state of the art for key Vision-to-Language (\V2L)
problems, such as image captioning and visual question answering.

This approach is reflected in many recent successful works on image captioning, such as~\cite{Chen2015CVPRMind,donahue2014long,karpathy2014deep,mao2014deep,vinyals2014show,yao2015describing,devlin2015language}.
Current state-of-the-art captioning methods use a CNN as an image `encoder' to produce a fixed-length vector representation~\cite{krizhevsky2012imagenet,lecun1998gradient,simonyan2014very,szegedy2014going}, which is then fed into the `decoder' RNN to generate a caption.

\begin{figure}[t!]
\centering
  \includegraphics[width=0.9\linewidth]{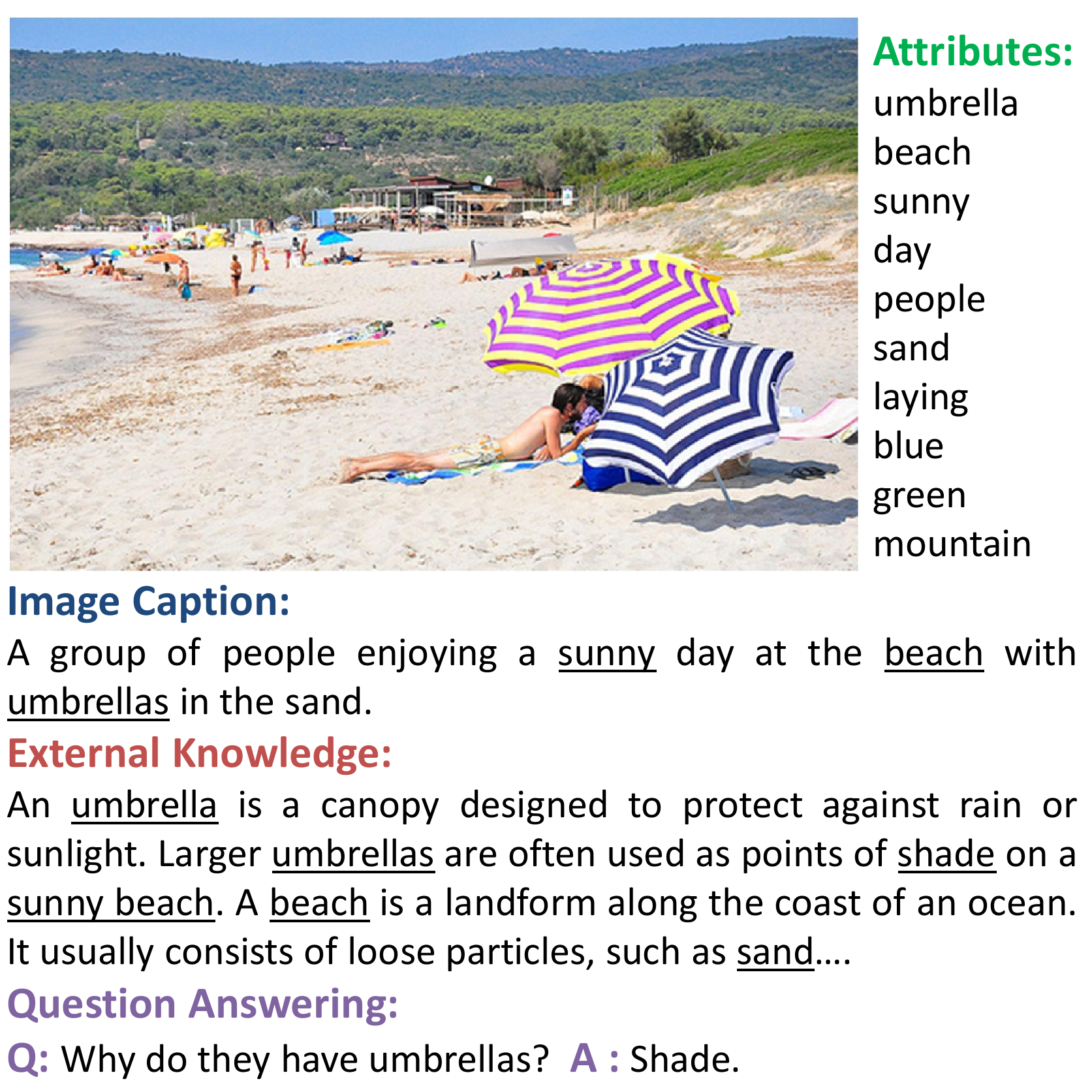}
  \vspace{-5pt}
  \caption{
    An example of the proposed \V2L system in action. Attributes are predicted by our CNN-based attribute prediction model.
    Image captions are generated by our attribute-based captioning generation model.
    All of the predicted attributes and generated captions, combined with the mined external knowledge
    from a large-scale knowledge base, are fed to an LSTM to produce the answer to the asked question.
    Underlined words indicate the information required to answer the question.}
  \label{img:example}
  \vspace{-15pt}
\end{figure}

Visual Question Answering (VQA) is a more recent challenge than image captioning. It is distinct from many problems in Computer Vision because the question to be answered is not determined until run time~\cite{antol2015vqa}. %
In this \V2L problem an image and a free-form, open-ended question about the image are presented to the method which is required to produce a suitable answer~\cite{antol2015vqa}. As in image captioning,
 the current state of the art in VQA~\cite{gao2015you,malinowski2015ask,ren2015image}
 relies on passing CNN features to an RNN language model.
 However, visual question answering is a significantly more complex problem than image captioning, not least because it requires accessing information not present in the image.  This may be common sense, or specific knowledge about the image subject. For example, given an image, such as Figure \ref{img:example}, showing `a group of people enjoying a sunny day at the beach with umbrellas', if one asks a question `why do they have umbrellas?', to answer this question, the machine must not only detect the scene `beach', but must know that `umbrellas are often used as points of shade on a sunny beach'. Recently, Antol \etal~\cite{antol2015vqa} also have suggested that VQA is a more ``AI-complete'' task since it requires multimodal knowledge beyond a single sub-domain. %

The contributions of this paper are two-fold.
First, we propose a {\em fully trainable} attribute-based neural network founded upon the CNN+RNN architecture,
that can be applied to multiple \textit{V2L} problems.
We do this by inserting an explicit representation of attributes of the scene which are meaningful to humans.
Each semantic attribute corresponds to a word mined from the training image descriptions, and represents higher-level
knowledge about the content of the image. A CNN-based classifier is trained for each attribute,
and the set of attribute likelihoods for an image form a high-level representation of image content.
An RNN is then trained to generate captions, or question answers, on the basis of the likelihoods.
Our attribute-based model yields significantly better performance than current state-of-the-art approaches in the task of image captioning. %

Based on the proposed attribute-based \textit{\V2L} model,
our second contribution is to introduce 
a method of incorporating knowledge external to the image, including common sense, into the VQA process.
In this work, we fuse the automatically generated description of an image with information extracted from an external
knowledge base (KB) to provide an answer to a general question about the image (See Figure \ref{frame}). The image description takes the form of a set of captions, and the external knowledge is text-based information mined from a Knowledge Base. Specifically, for each of the top-$k$ attributes detected in the image we generate a query which may be applied to a Resource Description Framework (RDF) KB, such as DBpedia.  RDF is the standard format for large KBs, of which there are many.  The queries are specified using Semantic Protocol And RDF Query Language (SPARQL). We encode the paragraphs extracted from the KB using Doc2Vec \cite{le2014distributed}, which maps paragraphs into a fixed-length feature representation. The encoded attributes, captions, and KB information are then input to an LSTM which is trained so as to maximise the likelihood of the ground truth answers in a training set. We further propose a question-guided knowledge selection scheme to improve the quality of  the extracted KB information. The knowledge that is
not related to the question is filtered out. The approach that we propose here combines the generality of information that using a KB allows with the generality of questions that the LSTM allows.
In addition, it achieves an accuracy of 70.98\% on the Toronto COCO-QA~\cite{ren2015image}, while the latest state of the art is 61.60\%. On the VQA~\cite{antol2015vqa} evaluation server (which does not publish ground truth answers for its test set), we also produce the state-of-the-art result, which is 59.50\%.

A preliminary version of this work was published at CVPR 2016 \cite{wu2015image,wu2015ask}. The new material in this paper comprises further experiments on two additional VQA datasets. More ablation models of the original model are implemented and studied. More importantly, a new model (A+C+Selected-K-LSTM) 
is introduced 
for the visual question answering task, leading to a new state-of-the-art result.

\section{Related work}
\label{sec:related_work}
\subsection{Attribute-based Representation} Using attribute-based models as a high-level representation has shown potential in many computer vision tasks such as object recognition, image annotation and image retrieval. Farhadi \etal \cite{farhadi2009describing} were
among the first to propose to use a set of visual semantic attributes to identify familiar objects, and to describe unfamiliar objects. %
Vogel and Schiele \cite{vogel2007semantic} used visual attributes describing scenes to characterize image regions and combined these local semantics into a global image description. Su \etal \cite{su2012improving} defined six groups of attributes to build intermediate-level features for image classification. Li \etal \cite{li2010object,li2012objects} introduced the concept of an `object bank' which enables objects to be used as attributes for scene representation.

\subsection{Image Captioning} The problem of annotating images with natural language at the scene level has long been studied in both computer vision and natural language processing. Hodosh \etal~\cite{hodosh2013framing} proposed to frame sentence-based image annotation as the task of ranking a given pool of captions. Similarly,~\cite{gong2014improving,jia2011learning,ordonez2011im2text} posed the task as a retrieval problem, but based on co-embedding of images and text in the same space. Recently, Socher \etal \cite{socher2014grounded} used neural networks to co-embed image and sentences together and Karpathy \etal~\cite{karpathy2014deep} co-embedded image crops and sub-sentences. %

Attributes have been used in many image captioning methods to fill the gaps in predetermined caption templates. Farhadi \etal~\cite{farhadi2010every}, for instance, used detections to infer a triplet of scene elements which is converted to text using a template. Li \etal \cite{li2011composing} composed image descriptions given computer vision based inputs such as detected objects, modifiers and locations using web-scale $n$-grams. Zhu \etal \cite{yao2010i2t} converted image parsing results into a semantic representation in the form of Web Ontology Language, which is converted to human readable text. A more sophisticated CRF-based method use of attribute detections beyond triplets was proposed by Kulkarni \textit{et al} \cite{kulkarni2013babytalk}. The advantage of template-based methods is that the resulting captions are more likely to be grammatically correct. The drawback is that they still rely on hard-coded visual concepts and suffer the implied limits on the variety of the output. %
Fang \etal \cite{fang2014captions} won the 2015 COCO Captioning Challenge with an approach that is similar to ours in as much as it applies a visual concept (i.e., attribute) detection process before generating sentences. They first learned $1000$ independent detectors for visual words based on a multi-instance learning framework and then used a maximum entropy language model conditioned on the set of visually detected words directly to generate captions. 

In contrast to the aforementioned two-stage methods, the recent dominant trend in \V2L is to use an architecture which connects a CNN to an RNN to learn the mapping from images to sentences directly. Mao \etal \cite{mao2014deep}, for instance, proposed a multimodal RNN (m-RNN) to estimate the probability distribution of the next word given previous words and the deep CNN feature of an image at each time step. Similarly, Kiros \etal \cite{kiros2014unifying} constructed a joint multimodal embedding space using a powerful deep CNN model and an LSTM that encodes text. Karpathy and Li \cite{Karpathy2014deepvs} also proposed a multimodal RNN generative model, but in contrast to \cite{mao2014deep}, their RNN is conditioned on the image information only at the first time step. Vinyals \etal \cite{vinyals2014show} combined deep CNNs for image classification with an LSTM for sequence modeling, to create a single network that generates descriptions of images. Chen \etal \cite{Chen2015CVPRMind} learn a bi-directional mapping between images and their sentence-based descriptions, which allows to reconstruct visual features given an image description. Xu \etal \cite{xu2015show} proposed a model based on visual attention. Jia \etal \cite{jia2015guilding} applied additional retrieved sentences to guide the LSTM in generating captions.

Interestingly, this end-to-end CNN-RNN approach ignores the image-to-word mapping which was an essential step in many of the previous image captioning systems detailed above~\cite{farhadi2010every,kulkarni2013babytalk,li2011composing,yang2011corpus}. The CNN-RNN approach has the advantage that it is able to generate a wider variety of captions, can be trained end-to-end, and outperforms the previous approach on the benchmarks. It is not clear, however, what the impact of bypassing the intermediate high-level representation is, and particularly to what extent the RNN language model might be compensating. Donahue \etal \cite{donahue2014long} described an experiment, for example, using tags and CRF models as a mid-layer representation for video to generate descriptions, but it was designed to prove that LSTM outperforms an SMT-based approach \cite{rohrbach2013translating}. It remains unclear whether the mid-layer representation or the LSTM leads to the success. Our paper provides several well-designed experiments to answer this question.

We thus here show not only a method for introducing a high-level representation into the CNN-RNN framework, and that doing so improves performance, but we also investigate the value of high-level information more broadly in \V2L tasks.  This is of critical importance at this time because \V2L has a long way to go, particularly in the generality of the images and text it is applicable to.

\begin{figure*}[t!]
  \centering
  \includegraphics[width=0.85\textwidth]{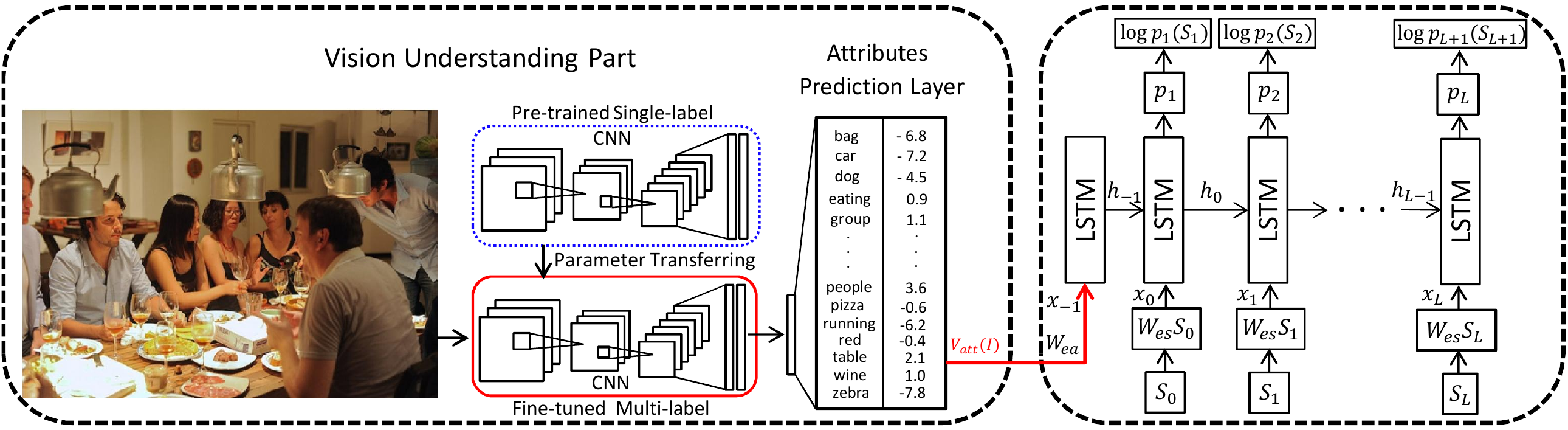}\\
  \vspace{-5pt}
  \caption{Our attribute-based image captioning framework. The image analysis module learns a mapping between an image and the semantic attributes through a CNN. The language module learns a mapping from the attributes vector to a sequence of words using an LSTM. }
  \label{img:cap_frame}
  \vspace{-5pt}
\end{figure*}

\subsection{Visual Question Answering} Malinowski \etal \cite{malinowski2014multi}
may be the first to study the VQA problem. They proposed a method that combines semantic parsing and image segmentation with a Bayesian approach to sampling from nearest neighbors in the training set. %
Tu \etal~\cite{tu2014joint} built a query answering system based on a joint parse graph from text and videos. Geman \etal~\cite{geman2015visual} proposed an automatic `query generator' that is trained on annotated images and produces a sequence of binary questions from any given test image. Each of these approaches places significant limitations on the form of question that can be answered.

Most recently, inspired by the significant progress achieved using deep neural network models in both computer vision and natural language processing, an architecture which combines a CNN and RNN to learn the mapping from images to sentences has become the dominant trend. Both Gao \etal~\cite{gao2015you} and Malinowski \etal \cite{malinowski2015ask} used RNNs to encode the question and output the answer.  Whereas Gao \etal~\cite{gao2015you} used two networks, a separate encoder and decoder, Malinowski \etal \cite{malinowski2015ask} used a single network for both encoding and decoding. Ren \etal \cite{ren2015image} focused on questions with a single-word answer and formulated the task as a classification problem using an LSTM. %
Antol \etal \cite{antol2015vqa} proposed a large-scale open-ended VQA dataset based on COCO, which is called VQA. %
Inspired by Xu \etal \cite{xu2015show} who encode visual attention in the Image Captioning, \cite{zhu2015visual7w,xu2015ask,Chen2015ABC,Jiang2015compositional,andreas2015deep,yang2015stacked} propose to use the spatial attention to help answering visual questions. \cite{xu2015ask, yang2015stacked, Noh2015image}
formulate the VQA as a classification problem and restrict the answer only can be drawn from a fixed answer space. %

Our framework also exploits both CNN and RNNs, but in contrast to preceding approaches which use only image features extracted from a CNN in answering a question, we employ multiple sources, including image content, generated image captions and mined external knowledge, to feed to an RNN to answer questions. Large-scale Knowledge Bases (KBs), such as Freebase~\cite{bollacker2008freebase} and DBpedia~\cite{auer2007dbpedia}, have been used successfully in several natural language Question Answering (QA) systems~\cite{berant2013semantic,ferrucci2010building}. However, VQA systems exploiting KBs are still relatively rare.

Zhu \etal \cite{zhu2015building} used a hand-crafted KB primarily containing image-related information such as category labels, attribute labels and affordance labels, but also some quantities relating to their specific question format such as GPS coordinates and similar. %
Instead of building a problem-specific KB, we use a pre-built large-scale KB (DBpedia \cite{auer2007dbpedia}) from which we extract information using a standard RDF query language. DBpedia has been created by extracting structured information from Wikipedia, and is thus significantly larger and more general than a hand-crafted KB. Rather than having a user pose their question in a formal query language, our VQA system is able to encode questions written in natural language automatically.  This is achieved without manually specified formalization, but rather depends on processing a suitable training set. The result is a model which is very general in the forms of question that it will accept.
The quality of the information in the KB is one of the primary issues in this approach to VQA. The problem is that KBs constructed by analysing Wikipedia and similar are patchy and inconsistent at best, and hand-curated KBs are inevitably very topic specific. Using visually-sourced information is a promising approach to solve this problem~\cite{Lin_2015_CVPR, Sadeghi_2015_CVPR}, but has a way to go before it might be usefully applied within our approach. After inspecting the database shows that the ‘comment’ field is the most generally informative about an attribute, as it contains a general text description of it. We therefore find this is still a feasible solution.

\section{Image Captioning using Attributes}
\label{sec:image_captioning}
Our image captioning model is summarized in Figure \ref{img:cap_frame}. The model includes an image analysis part and a caption generation part. In the image analysis part, we first use supervised learning to predict a set of attributes, based on words commonly found in image captions %
We solve this as a \textit{multi-label classification problem} and train a corresponding deep CNN by minimizing an element-wise logistic loss function. Secondly, a fixed length vector $\Att(I)$ is created for each image $I$, whose length is the size of the attribute set. Each dimension of the vector contains the prediction probability for a particular attribute. In the captioning generation part, we apply an LSTM-based sentence generator. In the baseline model, as in~\cite{gao2015you,ren2015image,vinyals2014show} we use a pre-trained CNN to extract image features $\CNN(I)$ which are fed into the LSTM directly. For the sake of completeness a fine-tuned version of this approach is also implemented.

\subsection{Attribute-based Image Representation}
\label{subsec:Attributes_Predictor}
Our first task is to describe the image content in terms of a set of attributes. An attribute vocabulary is first constructed. Unlike~\cite{kulkarni2013babytalk,yang2011corpus}, that use a vocabulary from separate hand-labeled training data, our semantic attributes are extracted from training captions and can be any part of speech, including object names (nouns), motions (verbs) or properties (adjectives). The direct use of captions guarantees that the most salient attributes for an image set are extracted. We use the $c$ ($c=256$) most common words in the training captions to determine the attribute vocabulary $\mathcal{V}_{att}$. Similar to \cite{fang2014captions}, the top $15$ most frequent closed-class words such as \texttt{`a',`on',`of'}  are removed since they are in nearly every caption. In contrast to~\cite{fang2014captions}, our vocabulary is not tense or plurality sensitive, for instance, \texttt{`ride'} and \texttt{`riding'} are classified as the same semantic attribute, similarly \texttt{`bag'} and \texttt{`bags'}. This significantly decreases the size of our attribute vocabulary. The full list of attributes can be found in the supplementary material. Our attributes represent a set of high-level semantic constructs, the totality of which the LSTM then attempts to represent in sentence form. Generating a sentence from a vector of attribute likelihoods exploits a much larger set of candidate words which are learned separately, allowing for greater flexibility in the generated text.

Given this attribute vocabulary, we can associate each image with a set of attributes according to its captions. We then wish to predict the attributes given a test image. Because we do not have ground truth bounding boxes for attributes, we cannot train a detector for each using the standard approach. Fang \etal~\cite{fang2014captions} solved a similar problem using a Multiple Instance Learning framework~\cite{zhang2005multiple} to detect visual words from images. Motivated by the relatively small number of times that each word appears in a caption, we instead treat this as a multi-label classification problem. To address the concern that some attributes may only apply to image sub-regions, we follow Wei \etal~\cite{wei2014cnn} in designing a region-based multi-label classification framework that takes an arbitrary number of sub-region proposals as input, then a shared CNN is associated with each proposal, and the CNN output results from different proposals are aggregated with max pooling to produce the final prediction.

\begin{figure}[t!]
  \centering
  \includegraphics[width=1\linewidth]{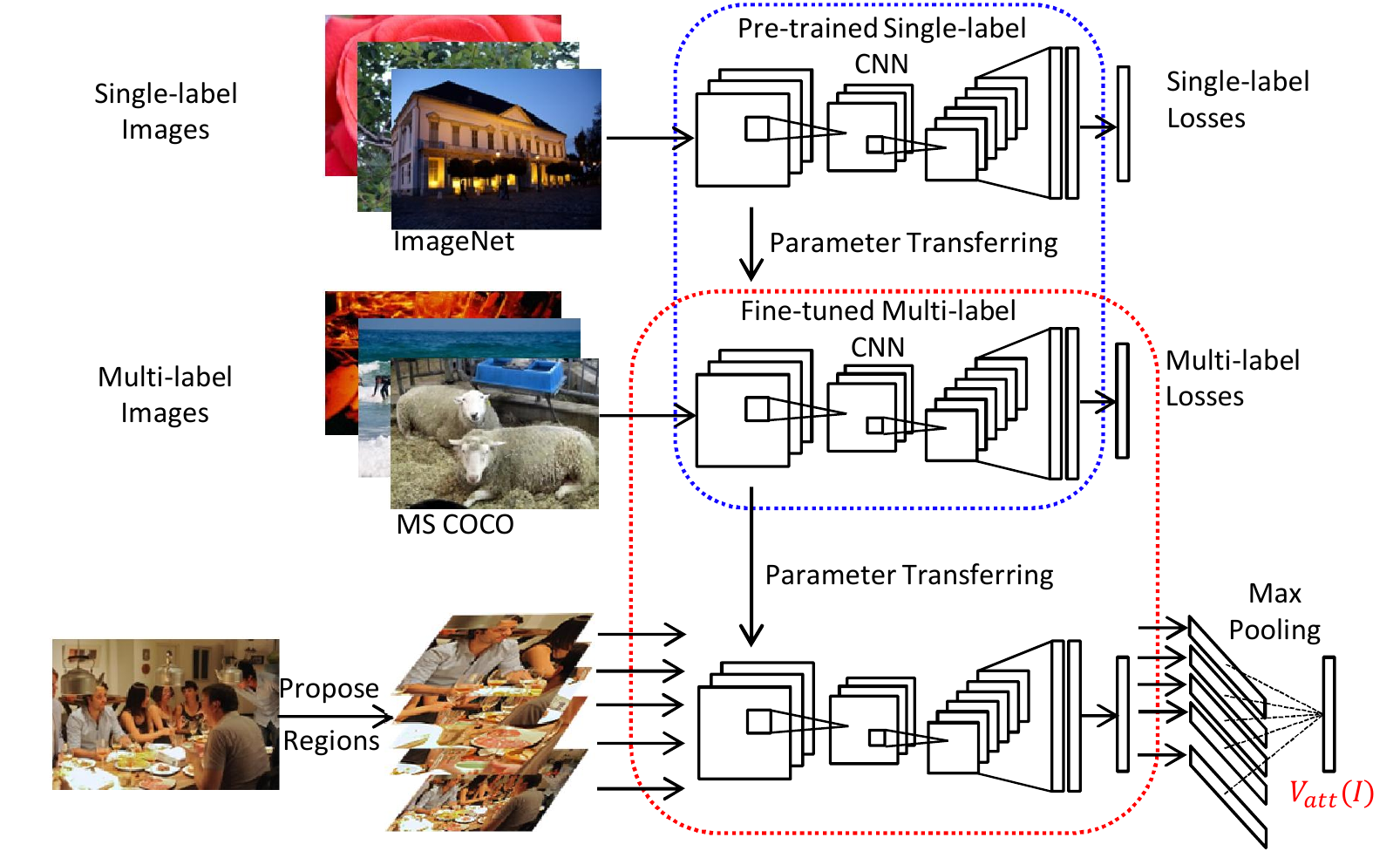}\\
  \caption{Attribute prediction CNN: the model is initialized from VggNet~\cite{simonyan2014very} pre-trained on ImageNet. The model is then fine-tuned on the target multi-label dataset. Given a test image, a set of proposal regions are selected and passed to the shared CNN, and finally the CNN outputs from different proposals are aggregated with max pooling to produce the final multi-label prediction, which gives us the high-level image representation, $\Att(I)$}
  \label{img:attributes}
  \vspace{-10pt}
\end{figure}

Figure \ref{img:attributes} summarizes the attribute prediction network. The model is a VggNet structure followed by a max-pooling operation on the regions with a multi-label loss. The CNN model is first initialized from the VggNet pre-trained on ImageNet.
The shared CNN is then fine-tuned on the target multi-label dataset (our image-attribute training data). In this step, the input is the global image and the output of the last fully-connected layer is fed into a $c$-way softmax over the $c$ class labels. The $c$ here represents the attributes vocabulary size. In contrast to \cite{wei2014cnn} who employs the squared loss, we find that element-wise logistic loss function performs better. Suppose that there are $N$ training examples and $\bm{y_i}=[y_{i1}, y_{i2},... , y_{ic}]$ is the label vector of the $i^{th}$ image, where $y_{ij}=1$ if the image is annotated with attribute $j$, and $y_{ij}=0$ otherwise. If the predictive probability vector is $\bm{p_i}=[p_{i1}, p_{i2},... , p_{ic}]$, the cost function to be minimized is
\begin{equation}
 J=\frac{1}{N}\sum_{i=1}^{N}\sum_{j=1}^{c}\log(1+\exp(-y_{ij}p_{ij}))
\end{equation}
During the fine-tuning process, the parameters of the last fully connected layer (i.e. the attribute prediction layer) are initialized with a Xavier initialization \cite{glorot2010understanding}. The learning rates of  `\texttt{fc6}' and `\texttt{fc7}' of the VggNet are initialized as 0.001 and the last fully connected layer is initialized as 0.01. All the other layers are fixed during training. We executed 40 epochs in total and decreased the learning rate to one tenth of the current rate for each layer after 10 epochs. The momentum is set to 0.9. The dropout rate is set to 0.5.

To predict attributes based on regions, we first extract hundreds of proposal windows from an image. However, considering the computational inefficiency of deep CNNs, the number of proposals processed needs to be small. Similar to \cite{wei2014cnn}, we first apply the normalized cut algorithm to group the proposal bounding boxes into $m$ clusters based on the IoU scores matrix. The top $k$ hypotheses in terms of the predictive scores reported by the proposal generation algorithm are kept and fed into the shared CNN. We also include the whole image in the hypothesis group. As a result, there are $mk+1$ hypotheses for each image. We set $m=10, k=5$ in all experiments. We use Multiscale Combinatorial Grouping (MCG)~\cite{PABMM2015} for the proposal generation. Finally, a cross hypothesis max-pooling is applied to integrate the outputs into a single prediction vector $\Att(I)$.

Since we formulate the attribute prediction as a multi-label problem, our attributes prediction network can be replaced by any other multi-label classification framework and it also can be benefit from the development of the multi-label classification researches. For example, to address the computational inefficiency of using a large numbers of proposed regions, we can apply an `R-CNN' architecture~\cite{girshick2015fast} so that we do not need to compute the convolutional feature map multiple times. The Regional Proposal Network~\cite{ren2015faster} can predict region proposal and attributes together so that we do not need the external region proposal tools. We even can consider the attributes dependencies by using the recently proposed CNN-RNN model \cite{wang2016cnn}. However, we leave them as the further work.

\subsection{Caption Generation Model}
\label{subsec:caption_model}
Similar to~\cite{Karpathy2014deepvs,mao2014deep,vinyals2014show}, we propose to train a caption generation model by maximizing the probability of the correct description given the image. However, rather than using image features directly as in typically the case, we use the semantic attribute prediction value $\Att(I)$ from the previous section as the input. Suppose that $\{S_1,...,S_L\}$ is a sequence of words. The log-likelihood of the words given their context words and the corresponding image can be written as:
\begin{equation}
    \log p(S|\Att(I))=\sum_{t=1}^L \log p(S_{t}|S_{1:t-1},\Att(I))
\end{equation}
where $p(S_t|S_{1:t-1},\Att(I))$ is the probability of generating the word $S_t$ given attribute vector $\Att(I)$ and previous words $S_{1:t-1}$. We employ the LSTM \cite{hochreiter1997long}, a particular form of RNN, to model this.

The LSTM is a memory cell encoding knowledge at every time step for what inputs have been observed up to this step. We follow the model used in \cite{zaremba2014learning}. Letting $\sigma$ be the sigmoid nonlinearity, the LSTM updates for time step $t$ given inputs $x_t$, $h_{t-1}$, $c_{t-1}$ are:

\vspace{-10pt}
\begin{eqnarray}
  i_t &=& \sigma(W_{xi}x_t+W_{hi}h_{t-1}+b_i) \\
  f_t &=& \sigma(W_{xf}x_t+W_{hf}h_{t-1}+b_f) \\
  o_t &=& \sigma(W_{xo}x_t+W_{ho}h_{t-1}+b_o) \\
  g_t &=& \tanh(W_{xc}x_t+W_{hc}h_{t-1}+b_c) \\
  c_t &=& f_t\odot c_{t-1}+i_t\odot g_t \\
  h_t &=& o_t \odot \tanh(c_t) \\
  p_{t+1} &=& {\rm softmax}(h_t)
\end{eqnarray}

Here, $i_t, f_t, c_t, o_t$ are the input, forget, memory, output state of the LSTM. The various $W$ matrices are trained parameters and $\odot$ represents the product with a gate value. $h_t$ is the hidden state at time step $t$ and is fed to a Softmax, which will produce a probability distribution $p_{t+1}$ over all words and indicate the word at time step $t+1$.

\vspace{3pt}
\noindent \textbf{Training details:} The LSTM model for image captioning is trained in an unrolled form. More formally, the LSTM takes the attributes vector $\Att(I)$ and a sequence of words $S=(S_0,...,S_L,S_{L+1})$, where $S_0$ is a special start word and $S_{L+1}$ is a special END token. Each word has been represented as a one-hot vector $S_t$ of dimension equal to the size of words dictionary. The words dictionaries are built based on words that occur at least 5 times in the training set, which lead to 2538, 7414, and 8791 words on Flickr8k, Flickr30k and MS COCO datasets separately. Note it is different from the semantic attributes vocabulary $\mathcal{V}_{att}$. The training procedure is as following: At time step $t=-1$, we set $x_{-1}=W_{ea}\Att(I)$, $h_{initial}=\vec{0}$ and $c_{initial}=\vec{0}$, where $W_{ea}$ is the learnable attributes embedding weights. This gives us an initial LSTM hidden state $h_{-1}$ which can be used in the next time step. From $t=0$ to $t=L$, we set $x_t=W_{es}S_t$ and the hidden state $h_{t-1}$ is given by the previous step, where $W_{es}$ is the learnable word embedding weights. The probability distribution $p_{t+1}$ over all words is then computed by the LSTM feed-forward process. Finally, on the last step when $S_{L+1}$ represents the last word, the target label is set to the END token.

\begin{figure}[t]
\begin{center}
   \includegraphics[width=0.95\linewidth]{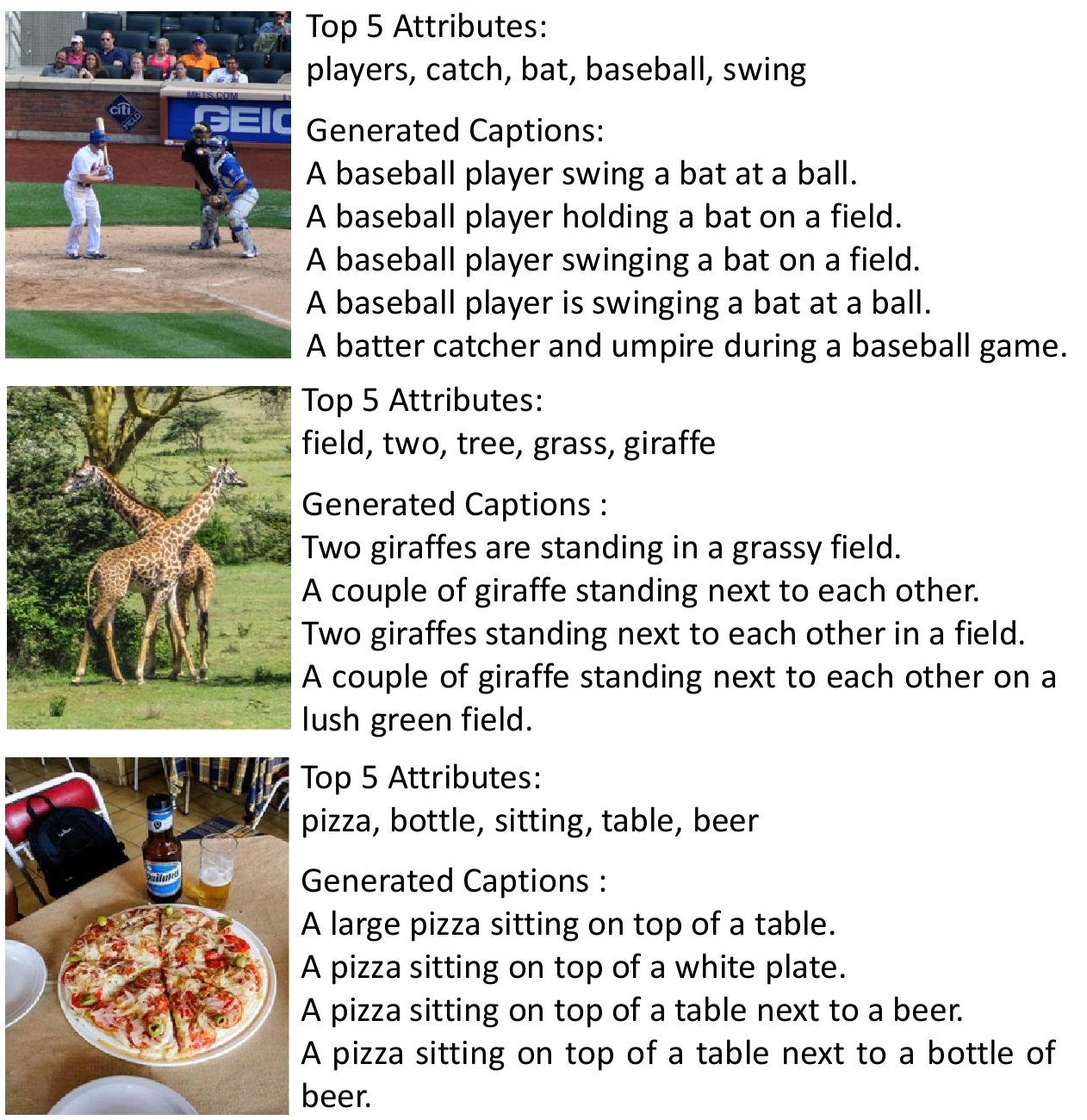}
\end{center}
\vspace{-10pt}
   \caption{Examples of predicted attributes and generated captions.}
   \vspace{-7pt}
   \label{att_caP_examples}
\end{figure}

Our training objective is to learn parameters $W_{ea}$, $W_{es}$ and all parameters in LSTM by minimizing the following cost function:
\begin{eqnarray}
    \mathcal{C}&=&-\frac{1}{N}\sum_{i=1}^N\log p(S^{(i)}|\Att(I^{(i)}))+\lambda_\ModParms\cdot||\ModParms||_2^2 \\
    &=&-\frac{1}{N}\sum_{i=1}^N\sum_{t=1}^{L^{(i)}+1}\log p_t(S_t^{(i)})+\lambda_\ModParms\cdot||\ModParms||_2^2
    \vspace{-3pt}
\end{eqnarray}
where $N$ is the number of training examples and $L^{(i)}$ is the length of the sentence for the $i$-th training example. $p_t(S_t^{(i)})$ corresponds to the activation of the Softmax layer in the LSTM model for the $i$-th input and $\ModParms$ represents model parameters, $\lambda_\ModParms\cdot||\ModParms||_2^2$ is a regularization term. We use SGD with mini-batches of 100 image-sentence pairs. The attributes embedding size, word embedding size and hidden state size are all set to 256 in all the experiments. The learning rate is set to 0.001 and clip gradients is 5. The dropout rate is set to 0.5.

To infer the sentence given an input image, we use Beam Search, \ie, we iteratively consider the set of $b$ best sentences up to time $t$ as candidates to generate sentences at time $t+1$, and only keep the best $b$ results. We set the $b$ to 5. Figure~\ref{att_caP_examples} shows some examples of the predicted attributes and generated captions. More results can be found in the supplementary material.

\section{A VQA Model with External Knowledge}
\label{sec:vqa}
\begin{figure*}[t!]
\begin{center}
\includegraphics[width=0.95\linewidth]{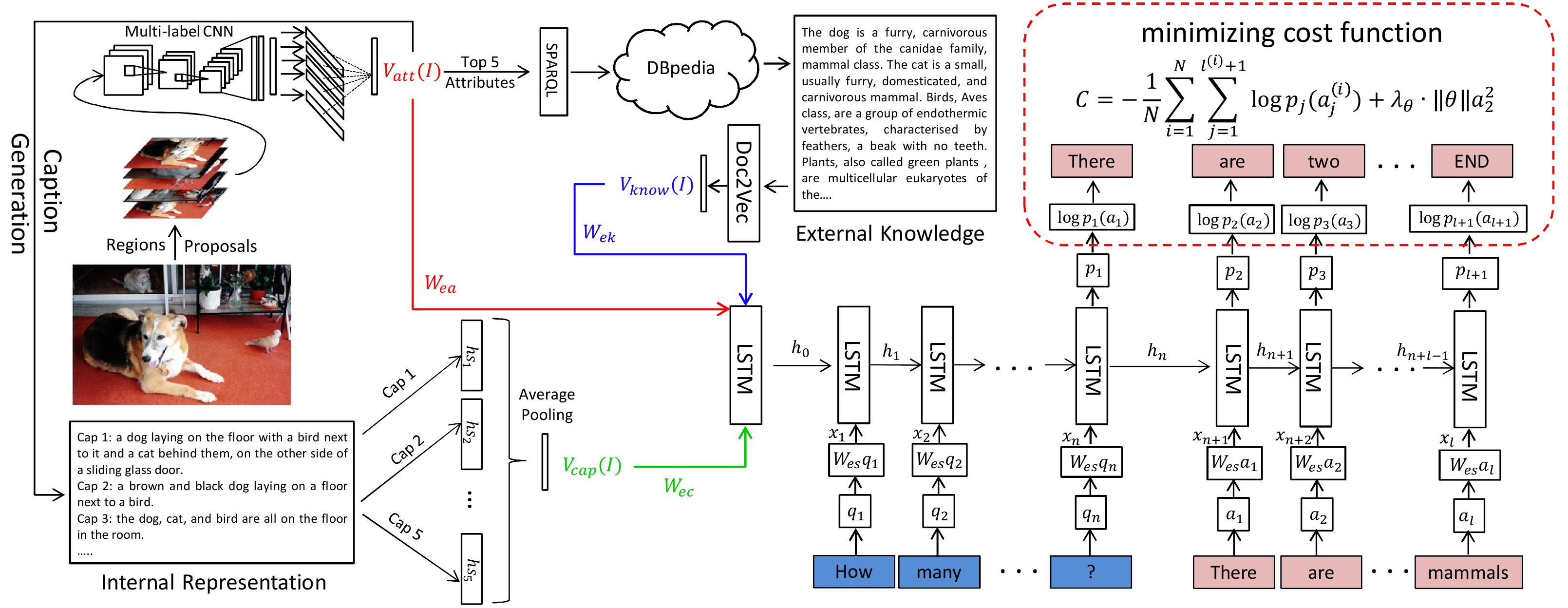}
\end{center}
\vspace{-10pt}
   \caption{Our proposed model: given an image, a CNN is first applied to produce the attribute-based representation \textcolor{red}{$\Att(I)$}. The internal textual representation is made up of image captions generated based on the image-attributes.
   The hidden state of the caption-LSTM after it has generated the last word in each caption is used as its vector representation. These vectors are then aggregated as \textcolor{green}{$\Cap(I)$} with average-pooling. The external knowledge is mined from the KB and the responses are encoded by Doc2Vec, which produces a vector \textcolor{blue}{$\Know(I)$}. The 3 vectors $\mathbf{V}$ are combined into a single representation of scene content, which is input to the VQA LSTM model that interprets the question and generates an answer.
   }
   \label{frame}
   \vspace{-10pt}
\end{figure*}

The key differentiator of our VQA model is that it is able to usefully combine image information with that extracted from a Knowledge Base, within the LSTM framework. The novelty lies in the fact that this is achieved by representing both of these disparate forms of information as text before combining them. Figure~\ref{frame} summarises how this is achieved: given an image, an attribute-based  representation $\Att(I)$ (in Section \ref{subsec:Attributes_Predictor}) is first generated and it will used as one of input sources of our VQA-LSTM model. The second input source are those captions generated in section \ref{subsec:caption_model}. Rather than inputing the generated words directly, the hidden state vector of the caption-LSTM after it has generated the last word in each caption is used to represent its content. Average-pooling is applied over the 5 hidden-state vectors, to obtain a vector representation $\Cap(I)$ for the image~$I$. The third input source is the textual knowledge which is mined from a large-scale knowledge base, the DBpedia. More details are shown in the following section.

\subsection{Relating to the Knowledge Base}

\begin{figure}[t]
  \centering
  \includegraphics[width=0.9\linewidth]{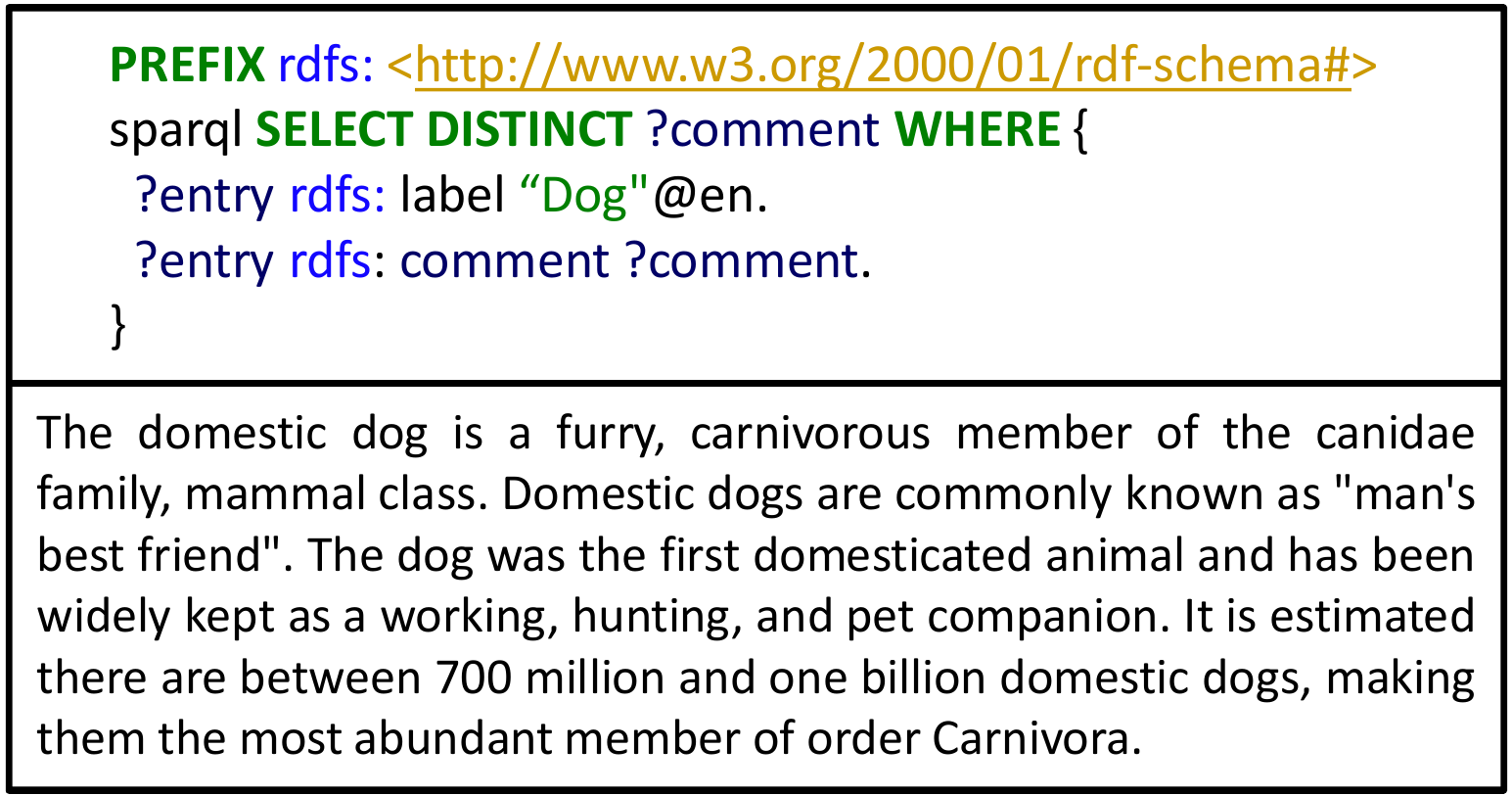}
  \vspace{-3pt}
  \caption{An example of SPARQL query language for the attribute \texttt{`dog'}. The mined text-based knowledge are shown below.}
  \label{img:example_rdf}
  \vspace{-13pt}
\end{figure}

The external data source that we use here is DBpeida~\cite{auer2007dbpedia} as a source of general background information, although any such KB could equally be applied. DBpeida is a structured database of information extracted from Wikipedia. The whole DBpedia dataset describes $4.58$ million entities, of which $4.22$ million are classified in a consistent ontology. The data can be accessed using an SQL-like query language for RDF called SPARQL. Given an image and its predicted attributes, we use the top-five\footnote{We only use top-5 attributes to query the KB because, based on observation of training data, an image typically contains 5-8 attributes. We also tested with top-10, but no improvements were observed.} most strongly predicted attributes to generate DBpedia queries. There are a range of problems with DBpedia and similar, however, including the sparsity of the information, and the inconsistency of its representation. Inspecting the database shows that the `comment' field is the most generally informative about an attribute, as it contains a general text description of it.  We therefore retrieve the comment text for each query term. The KB+SPARQL combination is very general, however, and could be applied problem specific KBs, or a database of common sense information, and can even perform basic inference over RDF. Figure~\ref{img:example_rdf} shows an example of the query language and returned text.

Since the text returned by the SPARQL query is typically much longer than the captions generated in the section \ref{subsec:caption_model}, we turn to Doc2Vec \cite{le2014distributed} to extract the semantic meanings\footnote{We investigated to use an LSTM to encode the mined paragraphs, but we observed little performance improvement, despite the additional training overhead.}. Doc2Vec, also known as Paragraph Vector, is an unsupervised algorithm that learns fixed-length feature representations from variable-length pieces of texts, such as sentences, paragraphs, and documents. Le \etal \cite{le2014distributed}  proved that it can capture the semantics of paragraphs. A Doc2Vec model is trained to predict words in the document given the context words. We collect 100,000 documents from  DBpedia to train a model with vector size 500. To obtain the knowledge vector $\Know(I)$ for image $I$, we combine the 5 returned paragraphs in to a single large paragraph, before semantic features using our pre-trained Doc2Vec model.

\vspace{-10pt}
\subsection{Question-guided Knowledge Selection}
\label{subsec:selected}
We incrementally implemented a question-guided knowledge selection scheme to rule out the noise information, since we observed that some mined knowledge are not necessary for answering the given question. For example, if the question is asking about the `dog' in the image, it does not make sense to input a piece of `bird' knowledge into the model, although the image does have a `bird' inside.

Given a question $Q$ and mined $n$ knowledge paragraphs using above KB+SPARQL combination, we first use our pre-trained Doc2Vec model to extract the semantic feature $V(Q)$ of the question and the feature $V(K_i)$ for each single knowledge paragraph, where $i \in n$. Then, we find the $k$ closest knowledge paragraphs to the question based on the cosine similarity between the $V(Q)$ and $V(K_i)$. Finally, we combine the $k$ selected knowledge paragraphs in to a single one and use the Doc2Vec model to extract its semantic feature. In our experiments, we set $n=10, k=5$.

\vspace{-10pt}
\subsection{An Answer Generation Model with Multiple Inputs}
We propose to train a VQA model by maximizing the probability of the correct answer given the image and question. We want our VQA model to be able to generate multiple word answers, so we formulate the answering process as a word sequence generation procedure. Let $Q=\{q_1,...,q_n\}$ represent the sequence of words in a question, and $A=\{a_1,...,a_l\}$ the answer sequence, where $n$ and $l$ are the length of question and answer, respectively. The log-likelihood of the generated answer can be written as: 
\begin{equation}
    \log p(A|I,Q)=\sum_{t=1}^l \log p(a_{t}|a_{1:t-1},I,Q)  
\end{equation}
where $p(a_t|a_{1:t-1},I,Q)$ is the probability of generating $a_t$ given image information $I$, question $Q$ and previous words $a_{1:t-1}$. We employ an encoder LSTM~\cite{hochreiter1997long} to take the semantic information from image $I$ and the question $Q$, while using a decoder LSTM to generate the answer. Weights are shared between the encoder and decoder LSTM.

In the training phase, the question $Q$ and answer $A$ are concatenated as $\{q_1,...,q_n,a_1,...,a_l,a_{l+1}\}$, where $a_{l+1}$ is a special END token. Each word is represented as a one-hot vector of dimension equal to the size of the word dictionary. The training procedure is as follows: at time step $t=0$, we set the LSTM input: 
\begin{equation}
   x_{initial}=[ W_{ea}\Att(I), W_{ec}\Cap(I), W_{ek}\Know(I) ] 
\end{equation}
where $W_{ea}$, $W_{ec}$, $W_{ek}$ are learnable embedding weights for the vector representation of attributes, captions and external knowledge, respectively. Given the randomly initialized hidden state, the encoder LSTM feeds forward to produce  hidden state $h_{0}$ which encodes all of the input information. From $t=1$ to $t=n$, we set $x_t=W_{es}q_t$ and the hidden state $h_{t-1}$ is given by the previous step, where $W_{es}$ is the learnable word embedding weights. The decoder LSTM runs from time step $n+1$ to $l+1$. Specifically, at time step $t=n+1$, the LSTM layer takes the input $x_{n+1}=W_{es}a_1$ and the hidden state $h_{n}$ corresponding to the last word of the question, where $a_1$ is the start word of the answer. The hidden state $h_{n}$ thus encodes all available information about the image and the question. The probability distribution $p_{t+1}$ over all answer words in the vocabulary is then computed by the LSTM feed-forward process. Finally, for the final step, when $a_{l+1}$ represents the last word of the answer, the target label is set to the END token.

Our training objective is to learn parameters $W_{ea}$, $W_{ec}$, $W_{ek}$, $W_{es}$ and all the parameters in the LSTM by minimizing the following cost function:
\begin{eqnarray}
    \mathcal{C}&=&-\frac{1}{N}\sum_{i=1}^N\log p(A^{(i)}|I,Q)+\lambda_\ModParms\cdot||\ModParms||_2^2 \\
    &=&-\frac{1}{N}\sum_{i=1}^N\sum_{j=1}^{l^{(i)}+1}\log p_j(a_j^{(i)})+\lambda_\ModParms\cdot||\ModParms||_2^2
\end{eqnarray}
where $N$ is the number of training examples, and $n^{(i)}$ and $l^{(i)}$ are the length of question and answer respectively for the $i$-th training example. Let $p_t(a_t^{(i)})$ correspond to the activation of the Softmax layer in the LSTM model for the $i$-th input and $\ModParms$ represent the model parameters. Note that $\lambda_\ModParms\cdot||\ModParms||_2^2$ is a regularization term, where $\lambda_\theta=0.5\times10^{-8}$. We use Stochastic gradient Descent (SGD) with mini-batches of 100 image-QA pairs. The attributes, internal textual representation, external knowledge embedding size, word embedding size and hidden state size are all 256 in all experiments. The learning rate is set to 0.001 and clip gradients is 5. The dropout rate is set to 0.5.

\section{Experiments}
\label{sec:exp}

\subsection{Evaluation on Image Captioning}
\subsubsection{Dataset}
We report image captioning results on the popular Flickr8k \cite{hodosh2013framing}, Flickr30k \cite{young2014image} and Microsoft COCO dataset \cite{lin2014microsoft}. These datasets contain 8,000, 31,000 and 123,287 images respectively, and each image is annotated with 5 sentences. In our reported results, we use pre-defined splits for Flickr8k.%
Because most of previous works in image captioning \cite{donahue2014long,fang2014captions,Karpathy2014deepvs,mao2014deep,vinyals2014show,xu2015show} are not evaluated on the official split for Flickr30k and MS COCO, for fair comparison, we report results with the widely used publicly available splits in the work of \cite{Karpathy2014deepvs}.%
We further tested on the actually MS COCO test set consisting of 40775 images (human captions for this split are not available publicly), and evaluated them on the COCO evaluation server.

\begin{table}[t]
\begin{center}
\scriptsize
\resizebox{1\linewidth}{!}{
 \begin{tabular}{ l c c c c|c }
    \multicolumn{6}{c}{Flickr8k}\\
    \Xhline{2\arrayrulewidth}
    \textbf{State-of-art-Flickr8k} & B-1 & B-2 & B-3 & B-4 & $\mathcal{PPL}$ \\ \hline
    Karpathy \& Li (NeuralTalk) \cite{Karpathy2014deepvs}& 0.58 & 0.38 & 0.25 & 0.16 & - \\
    Chen \& Zintick (Mind's Eye) \cite{Chen2015CVPRMind}&-&-&-&0.14&15.10 \\
    Google(NIC)\cite{vinyals2014show}& 0.66 & 0.42 & 0.27 &0.18&- \\
    Mao et al. (m-Rnn-AlexNet) \cite{mao2014deep}&0.57&0.39&0.26&0.17&24.39 \\
    Xu et al. (Hard-Attention) \cite{xu2015show}&0.67&0.46&0.31&0.21&- \\
    \Xhline{2\arrayrulewidth}
    \textbf{Baseline - \textit{CNN(I)}} & & & & & \\ \hline
    VggNet+LSTM & 0.56 & 0.37 & 0.24 & 0.16 & 15.71 \\
    VggNet-PCA+LSTM & 0.56 & 0.38 & 0.25 & 0.16 &16.07 \\
    VggNet+ft+LSTM & 0.64 & 0.43 & 0.30 & 0.20 & 14.69 \\
    \Xhline{2\arrayrulewidth}
    \textbf{Ours - $\Att(I)$} & & & & & \\ \hline
    \cellcolor[rgb]{0.7,0.7,0.7}{Att-GT+LSTM\textsuperscript{$\ddagger$}} & \cellcolor[rgb]{0.7,0.7,0.7}{0.76} & \cellcolor[rgb]{0.7,0.7,0.7}{0.57} & \cellcolor[rgb]{0.7,0.7,0.7}{0.41} & \cellcolor[rgb]{0.7,0.7,0.7}{0.29} & \cellcolor[rgb]{0.7,0.7,0.7}{12.52}\\
    Att-SVM+LSTM & 0.73 & 0.53 & \textbf{0.38} & 0.26 & 12.63\\
    Att-GlobalCNN+LSTM&0.72&0.53&\textbf{0.38}&\textbf{0.27}&12.63\\
    Att-RegionCNN+LSTM & \textbf{0.74} & \textbf{0.54} & \textbf{0.38} & \textbf{0.27} & \textbf{12.60}\\
    \Xhline{2\arrayrulewidth}
    \vspace{1pt}
 \end{tabular}}
 \scriptsize
 \resizebox{1\linewidth}{!}{
 \begin{tabular}{l c c c c|c }
    \multicolumn{6}{c}{Flickr30k}\\
    \Xhline{2\arrayrulewidth}
    \textbf{State-of-art-Flickr30k} & B-1 & B-2 & B-3 & B-4 & $\mathcal{PPL}$ \\ \hline
    Karpathy \& Li (NeuralTalk) \cite{Karpathy2014deepvs}& 0.57 & 0.37 & 0.24 & 0.16 & - \\
    Chen \& Zintick (Mind's Eye) \cite{Chen2015CVPRMind}& -&-&-&0.13&19.10 \\
    Google(NIC) \cite{vinyals2014show}& 0.66 & - & - &-&- \\
    Donahue et al. (LRCN) \cite{donahue2014long}&0.59&0.39&0.25&0.17&- \\
    Mao et al. (m-Rnn-AlexNet) \cite{mao2014deep}&0.54&0.36&0.23&0.15&35.11 \\
    Mao et al. (m-Rnn-VggNet) \cite{mao2014deep}&0.60&0.41&0.28&0.19&20.72 \\
    Xu et al. (Hard-Attention) \cite{xu2015show}&0.67&0.44&0.30&0.20&- \\
    \Xhline{2\arrayrulewidth}
    \textbf{Baseline - \textit{CNN(I)}} & & & & & \\ \hline
    VggNet+LSTM & 0.57 & 0.38 & 0.25 & 0.17 & 18.83 \\
    VggNet-PCA+LSTM & 0.59 & 0.40 & 0.26 & 0.17 & 18.92 \\
    VggNet+ft+LSTM & 0.67 & 0.47 & 0.31 & 0.21 & 16.62 \\\Xhline{2\arrayrulewidth}
    \textbf{Ours - $\Att(I)$} & & & & & \\ \hline
    \cellcolor[rgb]{0.7,0.7,0.7}{Att-GT+LSTM\textsuperscript{$\ddagger$}} & \cellcolor[rgb]{0.7,0.7,0.7}{0.78} & \cellcolor[rgb]{0.7,0.7,0.7}{0.57} & \cellcolor[rgb]{0.7,0.7,0.7}{0.42} & \cellcolor[rgb]{0.7,0.7,0.7}{0.30} & \cellcolor[rgb]{0.7,0.7,0.7}{14.88} \\
    Att-SVM+LSTM & 0.68 & 0.49 & 0.33 &0.23 & 16.01 \\
    Att-GlobalCNN+LSTM&0.70&0.50&0.35&0.27&16.00\\
    Att-RegionCNN+LSTM & \textbf{0.73} & \textbf{0.55} & \textbf{0.40} & \textbf{0.28} & \textbf{15.96}\\
    \Xhline{2\arrayrulewidth}
  \end{tabular}}
\end{center}
  \vspace{-5pt}
 \caption{BLEU-1,2,3,4 and $\mathcal{PPL}$ metrics compared to other state-of-the-art methods and our baseline on Flickr8k and Flickr30K dataset. $\ddagger$~indicates ground truth attributes labels are used, which (in \colorbox[rgb]{0.7,0.7,0.7}{gray}) will not participate in rankings. Our $\mathcal{PPL}s$ are based on Flickr8k and Flickr30k word dictionaries of size 2538 and 7414, respectively.}
 \vspace{-10pt}
\label{tab1} 
\end{table}

\subsubsection{Evaluation}
\textbf{Metrics:}
We report results with the frequently used BLEU metric \cite{papineni2002bleu} and sentence perplexity ($\mathcal{PPL}$). %
For MS COCO dataset, we additionally evaluate our model based on the metrics of METEOR \cite{banerjee2005meteor} and CIDEr \cite{vedantam2014cider}. %

\begin{table}[t]
\begin{center}
\scriptsize
  \resizebox{1\linewidth}{!}{%
  \begin{tabular}{ l c c c c c c|c}
    \Xhline{2\arrayrulewidth}
    \textbf{State-of-art} &B-1 & B-2 & B-3 & B-4 & M & C & $\mathcal{P}$ \\ \hline
    NeuralTalk \cite{Karpathy2014deepvs}& 0.63 & 0.45 & 0.32 & 0.23 & 0.20& 0.66&- \\
    Mind's Eye \cite{Chen2015CVPRMind}& -&-&-&0.19&0.20&-&11.60 \\
    NIC \cite{vinyals2014show}& - & - & - &0.28&0.24&0.86&- \\
    LRCN \cite{donahue2014long}&0.67&0.49&0.35&0.25&-&-&- \\
    Mao et al.\cite{mao2014deep}&0.67&0.49&0.34&0.24&-&-&13.60 \\
    Jia et al.\cite{jia2015guilding}&0.67&0.49&0.36&0.26&0.23&0.81&- \\
    MSR \cite{fang2014captions}&-&-&-&0.26&0.24&-&18.10\\
    Xu et al.\cite{xu2015show}&0.72&0.50&0.36&0.25&0.23&-&- \\
    Jin et al.\cite{jin2015aligning}&0.70&0.52&0.38&0.28&0.24&0.84&- \\
    \Xhline{2\arrayrulewidth}
    \textbf{Baseline-\textit{CNN(I)}} &  &  &  & & & & \\ \hline
    VNet+LSTM & 0.61 & 0.42 & 0.28 & 0.19 &0.19 &0.56&13.58  \\
    VNet-PCA+LSTM & 0.62 & 0.43 & 0.29 & 0.19 &0.20 &0.60&13.02 \\
    VNet+ft+LSTM & 0.68 & 0.50 & 0.37 & 0.25 & 0.22&0.73& 13.29 \\
    \Xhline{2\arrayrulewidth}
    \textbf{Ours-$\Att(I)$} & & & & & &&\\ \hline
    \cellcolor[rgb]{0.7,0.7,0.7}{Att-GT+LSTM\textsuperscript{$\ddagger$}} & \cellcolor[rgb]{0.7,0.7,0.7}{0.80} & \cellcolor[rgb]{0.7,0.7,0.7}{0.64} & \cellcolor[rgb]{0.7,0.7,0.7}{0.50} & \cellcolor[rgb]{0.7,0.7,0.7}{0.40} &\cellcolor[rgb]{0.7,0.7,0.7}{0.28} & \cellcolor[rgb]{0.7,0.7,0.7}{1.07}&\cellcolor[rgb]{0.7,0.7,0.7}{9.60} \\
    Att-SVM+LSTM & 0.69 & 0.52 & 0.38  &0.28 & 0.23& 0.82&12.62  \\
    Att-GlobalCNN+LSTM & 0.72  & 0.54  & 0.40   &0.30  & 0.25 & 0.83 &11.39    \\
    Att-RegionCNN+LSTM & \textbf{0.74} & \textbf{0.56} & \textbf{0.42}  & \textbf{0.31} & \textbf{0.26}& \textbf{0.94}& \textbf{10.49} \\ \hline
  \end{tabular}}
      \vspace{-4pt}
      \caption{BLEU-1,2,3,4, METEOR, CIDEr and $\mathcal{PPL}$ metrics compared to other state-of-the-art methods and our baseline on MS COCO dataset. $\ddagger$~indicates ground truth attributes labels are used, which (in \colorbox[rgb]{0.7,0.7,0.7}{gray}) will not participate in rankings. Our $\mathcal{PPL}s$ are based on MS COCO word dictionaries of size 8791.}
      \label{tab2}
      \vspace{-25pt}
\end{center}
\end{table}

\vspace{3pt}
\noindent\textbf{Baselines:}
To verify the effectiveness of our high-level attributes representation, we provide a baseline method. The baseline framework is same as the one proposed in section \ref{subsec:caption_model}, except that the attributes vector $\Att(I)$ is replaced by the last hidden layer of CNN directly. %
For the \textbf{VggNet+LSTM}, we use the second fully connected layer (\texttt{fc7}) as the image features, which has 4096 dimensions. In \textbf{VggNet-PCA+LSTM}, PCA is applied to decrease the feature dimension from 4096 to 1000. %
\textbf{VggNet+ft+LSTM} applies a VggNet that has been fine-tuned on the target dataset, based on the task of image-attributes classification.

\vspace{3pt}
\noindent\textbf{Our Approaches:}
We evaluate several variants of our approach: \textbf{Att-GT+LSTM} models use ground-truth attributes as the input while \textbf{Att-RegionCNN+LSTM} uses the attributes vector $\Att(I)$ predicted by the region based attributes prediction network in section \ref{subsec:Attributes_Predictor}. We also evaluate an approach \textbf{Att-SVM+LSTM} with linear SVM predicted attributes vector. %
We use the second fully connected layer of the fine-tuned VggNet to feed the SVM. To verify the effectiveness of the region based attributes prediction in the captioning task, the \textbf{Att-GlobalCNN+LSTM} is implemented by using the global image for attributes prediction.

\vspace{3pt}
\noindent\textbf{Results:} Table \ref{tab1} and \ref{tab2} report image captioning results on Flickr8k, Flickr30k and Microsoft COCO dataset. It is not surprising that \textbf{Att-GT+LSTM} model performs best, since ground truth attributes labels are used. We report these results here just to show the advances of adding an intermediate image-to-word mapping stage. Ideally, if we could train a perfectly accurate attribute predictor, we could obtain an outstanding improvement compared to both baseline and state-of-the-art methods. Indeed, apart from using ground truth attributes, our \textbf{Att-RegionCNN+LSTM} models generate the best results on all the three datasets over all evaluation metrics. Especially comparing with baselines, which do not contain an attributes prediction layer, our final models bring significant improvements, nearly 15\% for B-1 and 30\% for CIDEr on average. \textbf{VggNet+ft+LSTM} models perform better than other baselines because of the fine-tuning on the target dataset. However, they do not perform as well as our attributes-based models. \textbf{Att-SVM+LSTM} and \textbf{Att-GlobalCNN+LSTM} under-perform \textbf{Att-RegionCNN+LSTM}, indicating that region-based attributes prediction provides useful detail beyond whole image classification. Our final model also outperforms the current state-of-the-art listed in the tables. We also evaluated an approach (not shown in table) that combines CNN features and attributes vector together as the input of the LSTM, but we found this approach is not as good as using attributes vector only in the same setting.  In any case, above experiments show that an intermediate image-to-words stage (i.e. attributes prediction layer) bring us significant improvements.

\begin{table}[t!]
\begin{center}
\scriptsize
\resizebox{\linewidth}{!}{
  \begin{tabular}{ l c c c c c c c }
    \Xhline{2\arrayrulewidth}
    COCO-TEST & B-1 & B-2 & B-3&B-4 & M & R &CIDEr\\ \hline
    \textbf{5-Refs} &&&&&&& \\\hline
    Ours&0.73&0.56&0.41&0.31&0.25&0.53&0.92\\
    Human&0.66&0.47&0.32&0.22&0.2&0.48&0.85\\
    MSR \cite{fang2014captions}&0.70&0.53&0.39&0.29&0.25&0.52&0.91\\
    m-RNN \cite{mao2014deep}&0.68&0.51&0.37&0.27&0.23&0.50&0.79\\
    LRCN \cite{donahue2014long}&0.70&0.53&0.38&0.28&0.24&0.52&0.87\\
    Montreal \cite{xu2015show}&0.71&0.53&0.38&0.28&0.24&0.52&0.87\\
    Google \cite{vinyals2014show}&0.71&0.54&0.41&0.31&0.25&0.53&0.94\\
    NeuralTalk\cite{Karpathy2014deepvs}&0.65&0.46&0.32&0.22&0.21&0.48&0.67\\
    MSR Captivator\cite{devlin2015language}&0.72&0.54&0.41&0.31&0.25&0.53&0.93\\
    Nearest Neighbor \cite{devlin2015exploring}&0.70&0.52&0.38&0.28&0.24&0.51&0.89\\
    MLBL \cite{kiros2014multimodal}&0.67&0.50&0.36&0.26&0.22&0.50&0.74\\
    ATT \cite{you2016image}&0.73&0.57&0.42&0.32&0.25&0.54&0.94\\
    \Xhline{2\arrayrulewidth}
    \textbf{40-Refs} &&&&&&& \\\hline
    Ours&0.89&0.80&0.69&0.58&0.33&0.67&0.93\\
    Human&0.88&0.74&0.63&0.47&0.34&0.63&0.91\\
    MSR \cite{fang2014captions}&0.88&0.79&0.68&0.57&0.33&0.66&0.93\\
    m-RNN \cite{mao2014deep}&0.87&0.76&0.64&0.53&0.30&0.64&0.79\\
    LRCN \cite{donahue2014long}&0.87&0.77&0.65&0.53&0.32&0.66&0.89\\
    Montreal \cite{xu2015show}&0.88&0.78&0.66&0.54&0.32&0.65&0.89\\
    Google \cite{vinyals2014show}&0.89&0.80&0.69&0.59&0.35&0.68&0.95\\
    NeuralTalk\cite{Karpathy2014deepvs}&0.83&0.70&0.57&0.45&0.28&0.60&0.69\\
    MSR Captivator\cite{devlin2015language}&0.90&0.82&0.71&0.60&0.34&0.68&0.94\\
    Nearest Neighbor \cite{devlin2015exploring}&0.87&0.77&0.66&0.54&0.32&0.65&0.92\\
    MLBL \cite{kiros2014multimodal}&0.85&0.75&0.63&0.52&0.29&0.64&0.75\\
    ATT \cite{you2016image}&0.90&0.82&0.71&0.60&0.34&0.68&0.96\\
    \Xhline{2\arrayrulewidth}
  \end{tabular}}
  \vspace{-4pt}
  \caption{COCO evaluation server results. M and R stands for METEOR and ROUGE-L.
  Results using 5 references and 40 references captions are both shown. We only list the comparison results that have been officially published in the corresponding references. Please note some of them are concurrent results with this submission, such as \cite{you2016image}.}
  \label{tab3}
  \vspace{-23pt}
\end{center}
\end{table}

We further generated captions for the images in the COCO test set containing 40,775 images and evaluated them on the COCO evaluation server. These results are shown in Table \ref{tab3}. We achieve 0.73 on B-1, and surpass human performances on 13 of the 14 metrics reported. Other state-of-the-art methods are also shown for comparison.

\noindent\textbf{Human Evaluation:} We additionally perform a human evaluation on our proposed model, to evaluate the caption generation ability. We randomly sample 1000 results from the COCO validation dataset, generated by our proposed model Att-RegionCNN+LSTM and the baseline model VggNet+LSTM. Following the human evaluation protocol of the MS COCO Captioning Challenge 2015, two evaluation metrics are applied. M1 is the percentage of captions that are evaluated as better or equal to human caption and M2 is the percentage of captions that pass the Turing Test. Table \ref{tab_human} summarizes the human evaluation results. We can see our model outperforms the baseline model on both metrics. We did not evaluate on the test split because the human ground truth is not publicly available.

\begin{table}[h!]
\begin{center}
\vspace{-5pt}
\resizebox{\linewidth}{!}{
  \begin{tabular}{ >{\centering\arraybackslash}m{4cm} >{\centering\arraybackslash}m{2cm} >{\centering\arraybackslash}m{2cm} }
    \Xhline{2\arrayrulewidth}
     & Ours & VggNet+LSTM\\ \hline
    M1: percentage of captions that are evaluated as better or equal to human caption. &0.25&0.15\\ \hline
    M2: percentage of captions that pass the Turing Test. &0.30&0.19\\
    \Xhline{2\arrayrulewidth}
  \end{tabular}}
  \vspace{-3pt}
  \caption{ Human Evaluation on 1000 sampled results from MS COCO validation split.
  }
  \vspace{-20pt}
  \label{tab_human}
\end{center}
\end{table}

\begin{table}[t!]
\begin{center}
\scriptsize
\resizebox{\linewidth}{!}{
  \begin{tabular}{ c c c c c c}
    \Xhline{2\arrayrulewidth}
      & Ours & NIC\cite{vinyals2014show} & LRCN\cite{donahue2014long} & m-RNN\cite{mao2014deep} & NeuralTalk\cite{Karpathy2014deepvs}\\ \hline
    VIS Input Dim&256&1000&1000&4096&4096\\
    RNN Dim&256&512&1000$\times2/4$&256&300-600\\
    \Xhline{2\arrayrulewidth}
  \end{tabular}}
  \vspace{-3pt}
  \caption{Visual feature input dimension and properties of RNN. Our visual features has been encoded as a 256-d attributes score vector while other models need higher dimensional features to feed to RNN. According to the unit size of RNN, we achieve state-of-the-art using a relatively small dimensional recurrent layer. 
  }
  \vspace{-15pt}
  \label{tab3-2}
\end{center}
\end{table}

Table \ref{tab3-2} summarizes some properties of recurrent layers employed in some recent RNN-based methods.  We achieve state-of-the-art using a relatively low dimensional visual input feature and recurrent layer. Lower dimension of visual input and RNN normally means less parameters in the RNN training stage, as well as lower computation cost.

\subsection{Evaluation on Visual Question Answering}
We evaluate our model on four recent publicly available visual question answering datasets.
DAQURA-ALL is proposed in \cite{malinowski2014towards}. There are 7,795 training questions and 5,673 test questions. %
DAQURA-REDUCED is a reduced version of DAQURA-ALL. There are 3,876 training questions and only 297 test questions. This dataset is constrained to 37 object categories and uses only 25 test images. Two large-scale VQA data are constructed both based on MS COCO images. The Toronto COCO-QA Dataset \cite{ren2015image} contains 78,736 training and 38,948 testing examples, which are generated from 117,684 images. %
All of the question-answer pairs in this dataset are automatically converted from human-sourced image descriptions. Another benchmarked dataset is VQA~\cite{antol2015vqa}, which is a much larger dataset and contains 614,163 questions and 6,141,630 answers based on 204,721 MS COCO images. %
We randomly choose 5000 images from the validation set as our val set, with the remainder testing. The human ground truth answers for the actual VQA test split are not available publicly and only can be evaluated via the VQA evaluation server. Hence, we also apply our final model on a test split and report the overall accuracy. Table \ref{tab:data} displays some dataset statistics. %

\begin{table}[h]
\begin{center}
\scriptsize
\resizebox{\linewidth}{!}{
 \begin{tabular}{l c c c c}
 \Xhline{2\arrayrulewidth}
 &DAQURA&DAQURA&Toronto& \\ 
 &All&Reduced&COCO-QA&VQA \\\hline
 \# Images&1,449&1,423&117,684&204,721 \\
 \# Questions&12,468&4,173&117,684&614,163 \\
 \# Question Types&3&3&4&more than 20 \\
 \# Ans per Que&1&1&1&10\\
 \# Words per Ans&1+&1+&1&1+\\ \Xhline{2\arrayrulewidth}
 \end{tabular}}
 \vspace{-3pt}
 \caption{Some statistics about the DAQURA, Toronto COCO-QA Dataset \cite{ren2015image} and VQA dataset~\cite{antol2015vqa}.}
 \label{tab:data}
\end{center}
\end{table}

\subsubsection{Results on DAQURA}
\textbf{Metrics:}
Following \cite{ma2015learning, ren2015image}, the accuracy value (the proportion of correctly answered test questions), and the Wu-Palmer similarity (WUPS) \cite{wu1994verbs} are used to measure performance. The WUPS calculates the similarity between two words based on the similarity between their common subsequence in the taxonomy tree. If the similarity between two words is greater than a threshold then the candidate answer is considered to be right. We report on thresholds 0.9 and 0.0, following~\cite{ma2015learning,ren2015image}.

\vspace{3pt}
\noindent\textbf{Evaluations:}
To illustrate the effectiveness of our model, we provide two baseline models and several state-of-the-art results in table \ref{tab:DAQURA_All} and \ref{tab:DAQURA-Reduced}. The \textbf{Baseline} method is implemented simply by connecting a CNN to an LSTM. The CNN is a pre-trained (on ImageNet) VggNet model from which we extract the coefficients of the last fully connected layer. We also implement a baseline model \textbf{VggNet+ft-LSTM}, which applies a vggNet that has been fine-tuned on the COCO dataset, based on the task of image-attributes classification. We also present results from a series of cut down versions of our approach for comparison. \textbf{Att-LSTM} uses only the semantic level attribute representation $\Att$ as the LSTM input. To evaluate the contribution of the internal textual representation and external knowledge for the question answering, we feed the image caption representation $\Cap$ and knowledge representation $\Know$ with the $\Att$ separately, producing two models, \textbf{Att+Cap-LSTM} and \textbf{Att+Know-LSTM}. We also tested the \textbf{Cap+Know-LSTM}, for the experiment completeness. \textbf{Att+Cap+Know-LSTM} combines all the available information. Our final model is the \textbf{A+C+Selected-K-LSTM}, which uses the selected knowledge information (see section \ref{subsec:selected}) as the input. \textbf{GUESS} \cite{ren2015image} simply selects the modal answer from the training set for each of 4 question types.  %
\textbf{VIS+BOW}~\cite{ren2015image} performs multinomial logistic regression based on image features and a BOW vector obtained by summing all the word vectors of the question. \textbf{VIS+LSTM}~\cite{ren2015image} has one LSTM to encode the image and question, while \textbf{2-VIS+BLSTM}~\cite{ren2015image} has two image feature inputs, at the start and the end. Malinowski\etal~\cite{malinowski2015ask} propose a neural-based approach and Ma \textit{et al.}~\cite{ma2015learning} encodes both images and questions with a CNN. Yang~\etal~\cite{yang2015stacked} use a stacked attention networks to infer the answer progressively.

\begin{table}[t!]
\scriptsize
\begin{center}
\resizebox{\linewidth}{!}{
    \begin{tabular}{ l|c c c }
    \Xhline{2\arrayrulewidth}
    \textbf{DAQURA-All} &Acc(\%) & WUPS@0.9 & WUPS@0.0 \\ \hline
    Askneuron\cite{malinowski2015ask}&19.43&25.28&62.00\\
    Ma~\etal\cite{ma2015learning}&23.40&29.59&62.95\\
    Yang~\etal\cite{yang2015stacked}&\textbf{29.30}&35.10&68.60\\
    Noh~\etal\cite{Noh2015image}&28.98&34.80&67.81\\
    \Xhline{2\arrayrulewidth}
    \textbf{Baseline} &  &   &   \\ \hline
    VggNet-LSTM & 23.13 & 30.01 & 63.61 \\
    VggNet+ft-LSTM & 23.75 & 30.22 & 63.66 \\
    Human Baseline\cite{malinowski2015ask} & 50.20 & 50.82 & 62.27 \\
    \Xhline{2\arrayrulewidth}
    \textbf{Our-Proposal} &  &   &   \\ \hline
    Att-LSTM& 24.27& 30.41& 62.29 \\
    Att+Cap-LSTM& 27.04& 33.40& 67.65 \\
    Att+Know-LSTM& 24.89& 31.27& 66.11 \\
    Cap+Know-LSTM& 23.91& 30.64& 65.01 \\
    Att+Cap+Know-LSTM& 29.16& 35.30& 68.66 \\
    A+C+Selected-K-LSTM& 29.23& \textbf{35.37}&\textbf{68.72}\\
    \Xhline{2\arrayrulewidth}
    \end{tabular}}
    \vspace{-3pt}
    \caption{Accuracy, WUPS metrics compared to other state-of-the-art methods and our baseline on DAQURA-All.}
    \label{tab:DAQURA_All}
    \vspace{-8pt}
\end{center}
\end{table}

\begin{table}[t!]
\scriptsize
\begin{center}
\resizebox{\linewidth}{!}{
    \begin{tabular}{ l|c c c }
    \Xhline{2\arrayrulewidth}
    \textbf{DAQURA-Reduced} &Acc(\%) & WUPS@0.9 & WUPS@0.0 \\ \hline
    GUESS\cite{ren2015image}&18.24&29.65&77.59 \\
    VIS+BOW\cite{ren2015image}&34.17&44.99&81.48 \\
    VIS+LSTM\cite{ren2015image}&34.41&46.05&82.23 \\
    2-VIS+BLSTM\cite{ren2015image}&35.78&46.83&82.15 \\
    Askneuron\cite{malinowski2015ask}&34.68&40.76&79.54\\
    Ma~\etal\cite{ma2015learning}&39.66&44.86&83.06\\
    Xu~\etal\cite{xu2015ask}&40.07&-&-\\
    Yang~\etal\cite{yang2015stacked}&45.50&50.20&83.60\\
    Noh~\etal\cite{Noh2015image}&44.48&49.56&\textbf{83.95}\\
    \Xhline{2\arrayrulewidth}
    \textbf{Baseline} &  &   &   \\ \hline
    VggNet-LSTM & 38.72 & 43.97 & 83.01 \\
    VggNet+ft-LSTM & 39.13 & 44.03 & 83.33 \\
    Human Baseline\cite{malinowski2015ask} & 60.27 & 61.04 & 78.96 \\
    \Xhline{2\arrayrulewidth}
    \textbf{Our-Proposal} &  &   &   \\ \hline
    Att-LSTM& 40.07& 45.43& 82.67 \\
    Att+Cap-LSTM&44.78&50.07&83.85\\
    Att+Know-LSTM& 41.08& 46.04& 82.39 \\   
    Cap+Know-LSTM& 40.81& 45.04& 82.01 \\   
    Att+Cap+Know-LSTM& 45.79& 51.53&83.91\\
    A+C+Selected-K-LSTM& \textbf{46.13}& \textbf{51.83}&\textbf{83.95}\\
     \Xhline{2\arrayrulewidth}
    \end{tabular}}
    \vspace{-3pt}
    \caption{Accuracy, WUPS metrics compared to other state-of-the-art methods and our baseline on DAQURA-Reduced.}
    \label{tab:DAQURA-Reduced}
    \vspace{-20pt}
\end{center}
\end{table}

All of our proposed models outperform the \textbf{Baseline} method. And our final model \textbf{A+C+Selected-K-LSTM} achieves the best state-of-the-art on the DAQURA-Reduced set. \textbf{Att+Cap+Know-LSTM} performs not as good as \textbf{A+C+Selected-K-LSTM}, which shows the effectiveness of our question-based knowledge selection scheme.

\subsubsection{Results on Toronto COCO-QA}

\begin{table}[t!]
\scriptsize
\begin{center}
\resizebox{\linewidth}{!}{
    \begin{tabular}{ l|c c c }
    \Xhline{2\arrayrulewidth}
    \textbf{Toronto COCO-QA} &Acc(\%) & WUPS@0.9 & WUPS@0.0 \\ \hline
    GUESS\cite{ren2015image}&6.65&17.42&73.44 \\
    VIS+BOW\cite{ren2015image}&55.92&66.78&88.99 \\
    VIS+LSTM\cite{ren2015image}&53.31&63.91&88.25 \\
    2-VIS+BLSTM\cite{ren2015image}&55.09&65.34&88.64 \\
    Ma~\etal\cite{ma2015learning}&54.95&65.36&88.58\\
    Chen~\etal\cite{Chen2015ABC}&58.10&68.44&89.85\\
    Yang~\etal\cite{yang2015stacked}&61.60&71.60&90.90\\
    Noh~\etal\cite{Noh2015image}&61.19&70.84&90.61\\
    \Xhline{2\arrayrulewidth}
    \textbf{Baseline} &  &   &   \\ \hline
    VggNet-LSTM & 50.73 & 60.37 & 87.48 \\
    VggNet+ft-LSTM & 58.34 & 67.32 & 89.13 \\
    \Xhline{2\arrayrulewidth}
    \textbf{Our-Proposal} &  &   &   \\ \hline
    Att-LSTM& 61.38& 71.15& 91.58 \\
    Att+Cap-LSTM& 69.02& 76.20& 92.38 \\
    Att+Know-LSTM& 63.07& 72.22& 90.84 \\
    Cap+Know-LSTM& 64.31& 73.31& 90.01 \\
    Att+Cap+Know-LSTM& 69.73& 77.14& 92.50 \\
    A+C+Selected-K-LSTM& \textbf{70.98}& \textbf{78.35}& \textbf{92.87} \\
     \Xhline{2\arrayrulewidth}
    \end{tabular}}
    \vspace{-3pt}
    \caption{Accuracy, WUPS metrics compared to other state-of-the-art methods and our baseline on Toronto COCO-QA dataset.}
    \label{tab:coco_QA}
    \vspace{-7pt}
\end{center}
\end{table}

\vspace{3pt}
\noindent\textbf{Evaluations:}
\begin{table}[t!]
\scriptsize
\resizebox{\linewidth}{!}{
    \begin{tabular}{ l|c c c c }
    \Xhline{2\arrayrulewidth}
    \textbf{Toronto COCO-QA} &Object & Number & Color & Location \\ \hline
    GUESS\cite{ren2015image}&2.11&35.84&13.87& 8.93\\
    VIS+BOW\cite{ren2015image}&58.66&44.10&51.96&49.39 \\
    VIS+LSTM\cite{ren2015image}&56.53&46.10&45.87&45.52 \\
    2-VIS+BLSTM\cite{ren2015image}&58.17&44.79&49.53&47.34 \\
    Chen~\etal\cite{Chen2015ABC}&62.46&45.70&46.81&53.67\\
    Yang~\etal\cite{yang2015stacked}&64.50&48.60&57.90&54.00\\
    \Xhline{2\arrayrulewidth}
    \textbf{Baseline} & &   & &  \\ \hline
    VggNet-LSTM &53.71&45.37&36.23&46.37 \\
    VggNet+ft-LSTM &61.67&50.04&52.16&54.40 \\
    \Xhline{2\arrayrulewidth}
    \textbf{Our-Proposal} &  &   & &  \\ \hline
    Att-LSTM&63.92&51.83&57.29&54.84 \\
    Att+Cap-LSTM&71.30&69.98&61.50&60.98 \\
    Att+Know-LSTM&64.57&54.37&62.79&56.98 \\
    Cap+Know-LSTM& 65.61&55.13&62.02&57.28 \\
    Att+Cap+Know-LSTM& 71.45& \textbf{75.33}& \textbf{64.09}&60.98 \\
    A+C+Selected-K-LSTM& \textbf{73.66}&72.20& 62.97&\textbf{61.18} \\
     \Xhline{2\arrayrulewidth}
    \end{tabular}}
    \vspace{-3pt}
    \caption{Toronto COCO-QA accuracy (\%) per category.}
    \vspace{-15pt}
    \label{tab:coco_QA_cat}  
\end{table}
Table \ref{tab:coco_QA} reports the results on Toronto COCO-QA. All of our proposed models outperform the \textbf{Baseline} and all of the comparator state-of-the-art methods. Our final model \textbf{A+C+Selected-K-LSTM} achieves the best results. It surpasses the baseline by nearly 20\% and outperforms the previous state-of-the-art methods around 10\%. \textbf{Att+Cap-LSTM} clearly improves the results over the \textbf{Att-LSTM} model. This proves that internal textual representation plays a significant role in the VQA task. The \textbf{Att+Know-LSTM} model does not perform as well as \textbf{Att+Cap-LSTM} , which suggests that the information extracted from captions is more valuable than that extracted from the KB.  \textbf{Cap+Know-LSTM} also performs better than \textbf{Att+Know-LSTM}. This is not surprising because the Toronto COCO-QA questions were generated automatically from the MS COCO captions, and thus the fact that they can be answered by training on the captions is to be expected.  This generation process also leads to questions which require little external information to answer. The comparison on the Toronto COCO-QA thus provides an important benchmark against related methods, but does not really test the ability of our method to incorporate extra information.  It is thus interesting that the additional external information provides any benefit at all.

Table \ref{tab:coco_QA_cat} shows the per-category accuracy for different models. Surprisingly, the counting ability (see question type `Number') increases when both captions and external knowledge are included. This may be because some `counting' questions are not framed in terms of the labels used in the MS COCO captions. Ren \etal also observed similar cases. In \cite{ren2015image} they mentioned that ``there was some observable counting ability in very clean images with a single object type but the ability was fairly weak when different object types are present''. We also find there is a slight increase for the `color' questions when the KB is used. Indeed, some questions like `What is the color of the stop sign?' can be answered directly from the KB, without the visual cue. 

\subsubsection{Results on VQA}
Antol~\etal~\cite{antol2015vqa} provide the VQA dataset which is intended to support ``free-form and open-ended Visual Question Answering''. They also provide a metric for measuring performance: $\min\{\frac{\text{\# humans that said answer}}{3},1\}$ thus $100\%$ means that at least 3 of the 10 humans who answered the question gave the same answer. %

\vspace{3pt}
\noindent\textbf{Evaluation:} There are several splits for VQA dataset, such as the validation set, test-develop and test-standard set. We first tested several aspects of our models on the validation set (we randomly choose 5000 images from the validation set as our val set, with the remainder testing). %

Inspecting Table~\ref{tab:COCO_Results}, results the on VQA validation set, we see that the attribute-based \textbf{Att-LSTM} is a significant improvement over our \textbf{VggNet+LSTM} baseline. We also evaluate another baseline, the \textbf{VggNet+ft+LSTM}, which uses the penultimate layer of the attributes prediction CNN (in Section \ref{subsec:Attributes_Predictor}) as the input to the LSTM. Its overall accuracy on the VQA is 50.01, which is still lower than our proposed models (detailed results of different question types are not shown in Table~\ref{tab:COCO_Results} due to the limited space.) Adding either image captions or external knowledge further improves the result. Our final model \textbf{A+C+S-K-LSTM} produces the best results, outperforming the baseline \textbf{VggNet-LSTM} by 11\%. %

\begin{figure}[b!]
\vspace{-10pt}
\begin{center}
   \includegraphics[width=0.9\linewidth]{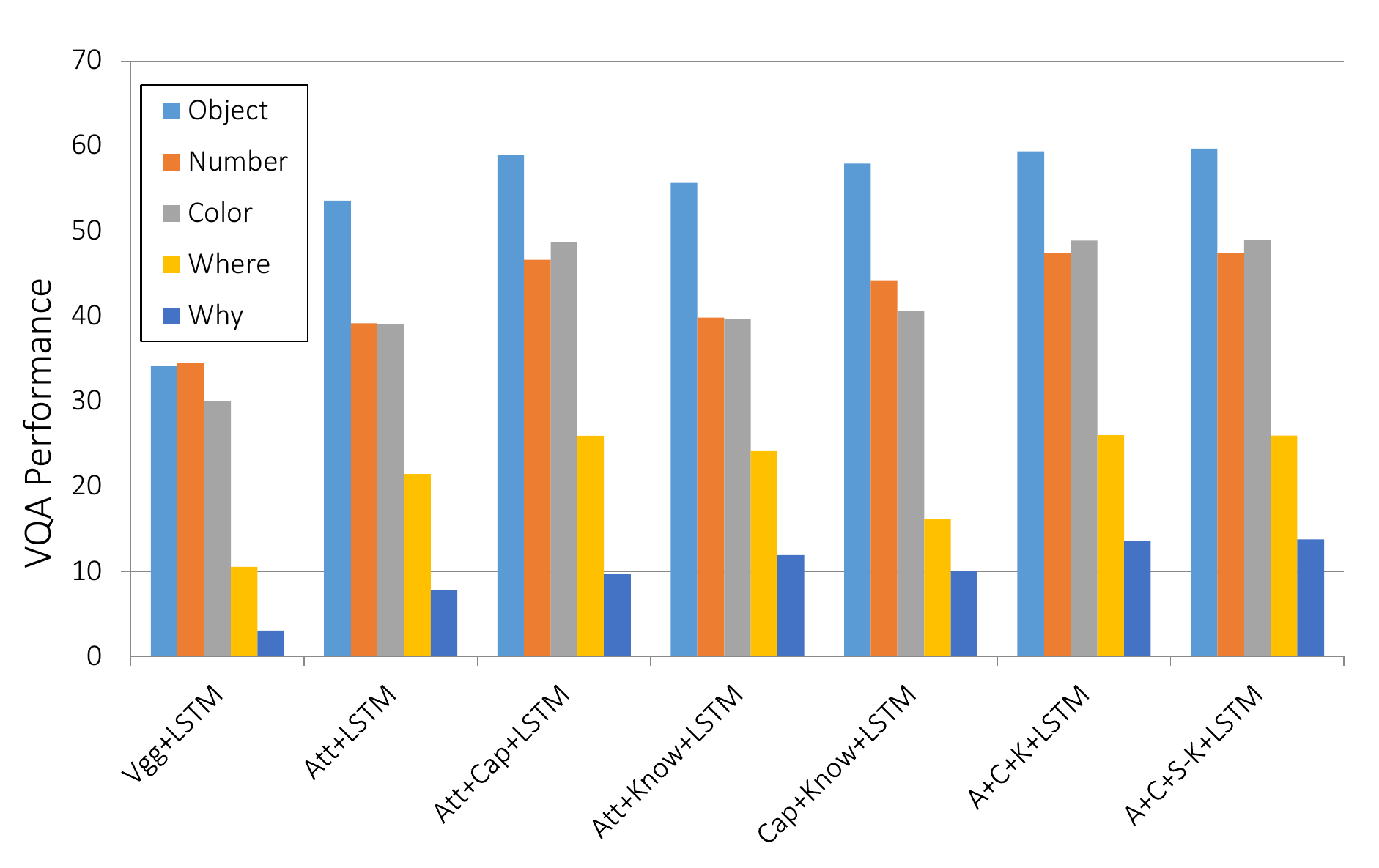}
\end{center}
\vspace{-13pt}
   \caption{Performance on five question categories for different models.}
\label{performance_trend}
\end{figure}

Figure~\ref{performance_trend} relates the performance of the various models on five categories of questions. The `object' category is the average of the accuracy of question types starting with `what kind/type/sport/animal/brand...', while the `number' and `color' category corresponds to the question type `how many' and `what color'. The performance comparison across categories is of particular interest here because answering different classes of questions requires different amounts of external knowledge.  The 'Where' questions, for instance, require knowledge of potential locations, and 'Why' questions typically require general knowledge about people's motivation. 'Number' and 'Color' questions, in contrast, can be answered directly. The results show that for 'Why' questions, adding the KB improves performance by more than $50\%$ (Att-LSTM achieves $7.77\%$ while Att+Know-LSTM achieves $11.88\%$), and that the combined A+C+K-LSTM achieves $13.53\%$. We further improve it to $13.76\%$ by using the question-guided knowledge selected model A+C+S-K-LSTM. Compared with the Att-LSTM model, the performance gain of the Cap+Know-LSTM model mainly come from the `why' and `where' started questions. This means that the external knowledge we employed in the model provide useful information to answer such questions. The figure \ref{img:example} shows an real example produced by our model. More questions that require common-sense knowledge to answer can be found in the supplementary materials.

\begin{table}[t!]
\centering
\resizebox{\linewidth}{!}{
 \begin{tabular}{lclccccc}
  \Xhline{2\arrayrulewidth}
  \multicolumn{1}{c}{}         & Our-Baseline &  & \multicolumn{5}{c}{Our Proposal}                                          \\ \cline{2-2} \cline{4-8}
  \multicolumn{1}{c}{Question} & VggNet       &  & Att              & Att+Cap          & Att+Know         & A+C+K            &A+C+S-K\\
  \multicolumn{1}{c}{Type}     & +            &  & +                & +                & +                & +                &+\\
  \multicolumn{1}{c}{}         & LSTM         &  & LSTM             & LSTM             & LSTM             & LSTM             &LSTM\\ \hline
  what is                      & 21.41        &  & 34.63            & 42.21            & 37.11            & \textbf{42.52}   &42.51\\
  what colour                  & 29.96        &  & 39.07            & 48.65            & 39.68            & 48.86            &\textbf{48.89}\\
  what kind                    & 24.15        &  & 41.22            & 47.93            & 46.16            & \textbf{48.05}   & 48.02 \\
  what are                     & 23.05        &  & 38.87            & 47.13            & 41.13            & 47.21            & \textbf{47.27} \\
  what type                    & 26.36        &  & 41.71            & 47.98            & 44.91            & 48.11            & \textbf{48.14} \\
  is the                       & 71.49        &  & 73.22            & 74.63            & 74.40            & \textbf{74.70}   & \textbf{74.70} \\
  is this                      & 73.00        &  & 75.26            & 76.08            & \textbf{76.56}   & 76.14            & 76.17 \\
  how many                     & 34.42        &  & 39.14            & 46.61            & 39.78            & \textbf{47.38}   & \textbf{47.38} \\
  are                          & 73.51        &  & 75.14            & 76.01            & 75.75            & 76.14            & \textbf{76.15} \\
  does                         & 76.51        &  & 76.71            & 78.07            & 76.55            & \textbf{78.11}   & \textbf{78.11} \\
  where                        & 10.54        &  & 21.42            & 25.92            & 24.13            & \textbf{26.00}   & 25.96 \\
  is there                     & 86.66        &  & 87.10            & 86.82            & 85.87            & 87.01            & \textbf{87.33} \\
  why                          & 3.04         &  & 7.77             & 9.63             & 11.88            & 13.53            & \textbf{13.76}  \\
  which                        & 31.28        &  & 36.60            & \textbf{39.55}   & 37.71            & 38.70            & 38.83 \\
  do                           & 76.44        &  & 75.76            & 78.18            & 75.25            & 78.42            & \textbf{78.44} \\
  what does                    & 15.45        &  & 19.33            & 21.80            & 19.50            & 22.16            & \textbf{22.71} \\
  what time                    & 13.11        &  & 15.34            & 15.44            & \textbf{15.47}   & 15.34            & 15.17 \\
  who                          & 17.07        &  & 22.56            & 25.71            & 21.23            & 25.74            & \textbf{25.97} \\
  what sport                   & 65.65        &  & 91.02            & 93.96            & 90.86            & \textbf{94.20}   & 94.18 \\
  what animal                  & 27.77        &  & 61.39            & 70.65            & 63.91            & 71.70            & \textbf{72.33} \\
  what brand                   & 26.73        &  & 32.25            & 33.78            & 32.44            & 34.60            & \textbf{35.68} \\
  others                       & 44.37        &  & 50.23            & 53.29            & 52.11            & 53.45            & \textbf{53.53}     \\ \hline
  Overall                      & 44.93        &  & 51.60            & 55.04            & 53.79            & 55.96            & \textbf{56.17} \\ \Xhline{2\arrayrulewidth}
 \end{tabular}}
\vspace{1pt}
\caption{Results on the open-answer task for various question types on VQA validation set. All results are in terms of the evaluation metric from the VQA evaluation tools. The overall accuracy for the model of \textbf{VggNet+ft+LSTM} and \textbf{Cap+Know+LSTM} is 50.01 and 52.31 respectively. Detailed results of different question types for these two models are not shown in the table due to the limited space.}
\vspace{-12pt}
\label{tab:COCO_Results}
\end{table}

We have also tested on the VQA test-dev and test-standard consisting of 60,864 and 244,302 questions (for which ground truth answers are not published) using our final A+C+S-K-LSTM model, and evaluated them on the VQA evaluation server. Table \ref{tab:server_results_2} shows the server reported results. The results on the Test-dev can be found in the supplementary material.

Antol \etal \cite{antol2015vqa} provide several results for this dataset. In each case they encode the image with the final hidden layer from VggNet, and questions and captions are encoded using a BOW representation. A softmax neural network classifier with 2 hidden layers and 1000 hidden units (dropout 0.5) in each layer with tanh non-linearity is then trained, the output space of which is the 1000 most frequent answers in the training set. They also provide an LSTM model followed by a softmax layer to generate the answer. Two version of this approach are used, one which is given only the question and the image, and one which is given only the question (see~\cite{antol2015vqa} for details). %
Our final model outperforms all the listed approaches according to the overall accuracy. Figure \ref{results_examples} provides some indicative results. More results can be found in the supplementary material.

\vspace{-8pt}
\section{Conclusions}

In this paper, we first examined the importance of introducing an intermediate attribute prediction layer into the predominant CNN-LSTM framework,
which was neglected by almost all previous work.
We implemented an attribute-based model which can be applied to the task of image captioning. We have shown that an explicit representation of image content improves \V2L performance, in all cases. Indeed, at the time of submitting this paper,
our image captioning model outperforms the state-of-the-art on several captioning datasets.

Secondly, in this paper we have shown that it is possible to extend the state-of-the-art RNN-based VQA approach so as to incorporate the large volumes of information required to answer general, open-ended, questions about images. The knowledge bases which are currently available do not contain much of the information which would be beneficial to this process, but nonetheless can still be used to significantly improve performance on questions requiring external knowledge (such as 'Why' questions).  The approach that we propose is very general, however, and will be applicable to more informative knowledge bases should they become available. We further implement a knowledge selection scheme which reflects both of the content of the question and the image, in order to extract more specifically related information. Currently our system is the state-of-the-art on three VQA datasets and produces the best results on the VQA evaluation server.

Further work includes generating knowledge-base queries which reflect the content of the question and the image, in order to extract more specifically related information. The Knowledge Base itself also can be improved. For instance, Open-IE provides more general common-sense knowledge such as `cats eat fish'. Such knowledge will help answer high-level questions.

\begin{table}[t!]
\centering
\scriptsize
\resizebox{\linewidth}{!}{
\begin{tabular}{lccclc}
\Xhline{2\arrayrulewidth}
\multicolumn{1}{c}{VQA}& \multicolumn{3}{c}{Answer Type} &  & Overall \\ \cline{2-4}
\multicolumn{1}{c}{Test-standard} & Yes/No & Other & Number &  &  \\ \hline
LSTM Q \cite{antol2015vqa}& 78.12 & 26.99 & 34.94 &  & 48.89 \\
LSTM Q+I \cite{antol2015vqa}& 79.01 & 36.80 & 35.55 &  & 54.06 \\
IBOWING \cite{zhou2015simple}&76.76&42.62&34.98&&55.89\\
NMN \cite{andreas2015deep}& 81.16 & 44.01 & 37.70 &  & 58.66\\
DNMN \cite{andreas2016learning}&80.98&45.81&37.48&&59.44\\
SMem \cite{xu2015ask}&80.80&43.48&37.53&&58.24\\
SAN \cite{yang2015stacked}& 79.11 & 46.42 &36.41  &  & 58.85\\
DDPnet\cite{Noh2015image}&80.28&42.24&36.92&&57.36\\
Human \cite{antol2015vqa}&95.77&72.67&83.39&&83.30\\
\hline
Ours & 81.10 & 45.90 & 37.18 &  & \textbf{59.50} \\ \Xhline{2\arrayrulewidth}
\end{tabular} }
\vspace{-3pt}
\caption{VQA Open-Ended evaluation server results. Accuracies for different answer types and overall performances on the test-standard. We only list the published results before this submission, the whole list of the leanding board can be found from http://www.visualqa.org/roe.html}
\vspace{-12pt}
\label{tab:server_results_2}
\end{table}

\vspace{-7pt}
\section*{Acknowledgements}
This research was in part supported by the Data to Decisions Cooperative Research Centre.
Correspondence should be addressed to C. Shen.

\begin{figure*}[t]
\resizebox{\linewidth}{!}{
\begin{tabular}{lcccc}
 \multicolumn{2}{c}{\includegraphics[height=3cm]{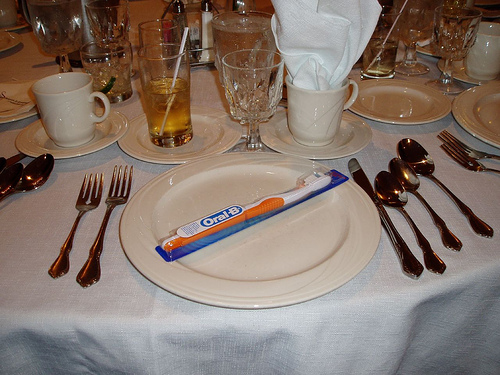}}&\includegraphics[height=3cm]{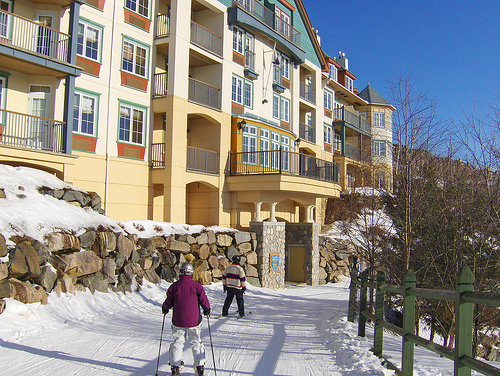}&\includegraphics[height=3cm]{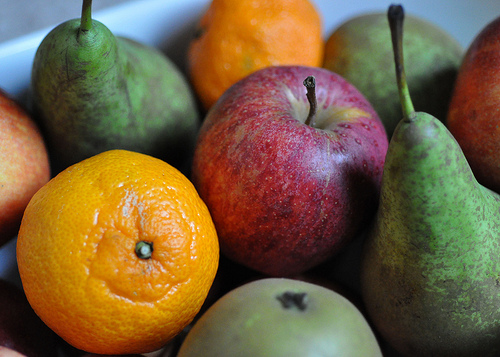}&\includegraphics[height=3cm]{img/results/008.jpg}\\
 \multicolumn{2}{c}{What color is the tablecloth?}&How many people in the photo?&What is the red fruit?&What are these people doing?\\ \hline
 \textit{Ours:} &\color{green}{white}&\color{green}{2}&\color{green}{apple}&\color{green}{eating}\\
 \textit{Vgg+LSTM:} &\color{red}{red}&\color{red}{1}&\color{red}{banana}&\color{red}{playing}\\
 \textit{Ground Truth:} &white&2&apple&eating\\ \Xhline{2\arrayrulewidth}
 \vspace{1pt}
\end{tabular}}
\resizebox{\linewidth}{!}{
\begin{tabular}{lcccc}
 \multicolumn{2}{c}{\includegraphics[height=3cm]{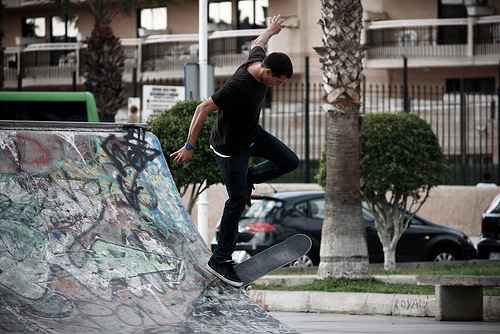}}&\includegraphics[height=3cm]{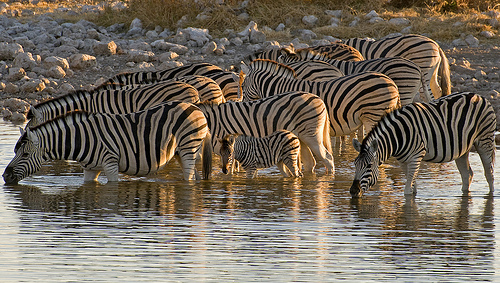}&\includegraphics[height=3cm]{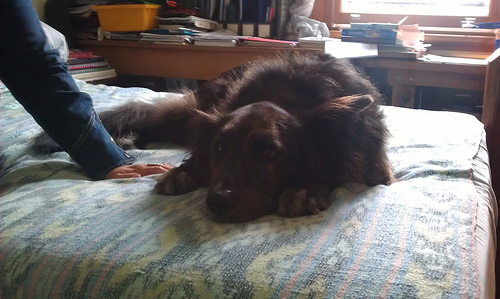}&\includegraphics[height=3cm]{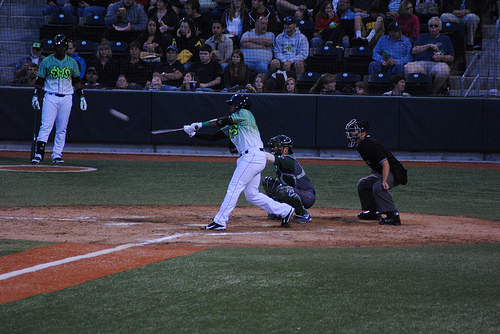}\\
 \multicolumn{2}{c}{Why are his hands outstretched?}&Why are the zebras in water?&Is the dog standing or laying down?&Which sport is this?\\ \hline
 \textit{Ours:} &\color{green}{balance}&\color{green}{drinking}&\color{green}{laying down}&\color{green}{baseball}\\
 \textit{Vgg+LSTM:} &\color{red}{play}&\color{red}{water}&\color{red}{sitting}&\color{red}{tennis}\\
 \textit{Ground Truth:} &balance&drinking&laying down&baseball\\ \Xhline{2\arrayrulewidth}
 \vspace{-3pt}
\end{tabular}}
\vspace{-7pt}
\caption{Some example cases where our final model gives the correct answer while the base line model \textbf{VggNet-LSTM} generates the wrong answer. All results are from the VQA dataset. More results can be found in the supplementary material.}
\label{results_examples}
\vspace{-15pt}
\end{figure*}
\ifCLASSOPTIONcaptionsoff
  \newpage
\fi

\bibliographystyle{IEEEtran}
\bibliography{IEEEabrv,./myref}

\vspace{-1cm}
\begin{wrapfigure}{l}{7em}
\vspace{-10pt}
  \includegraphics[width=7em]{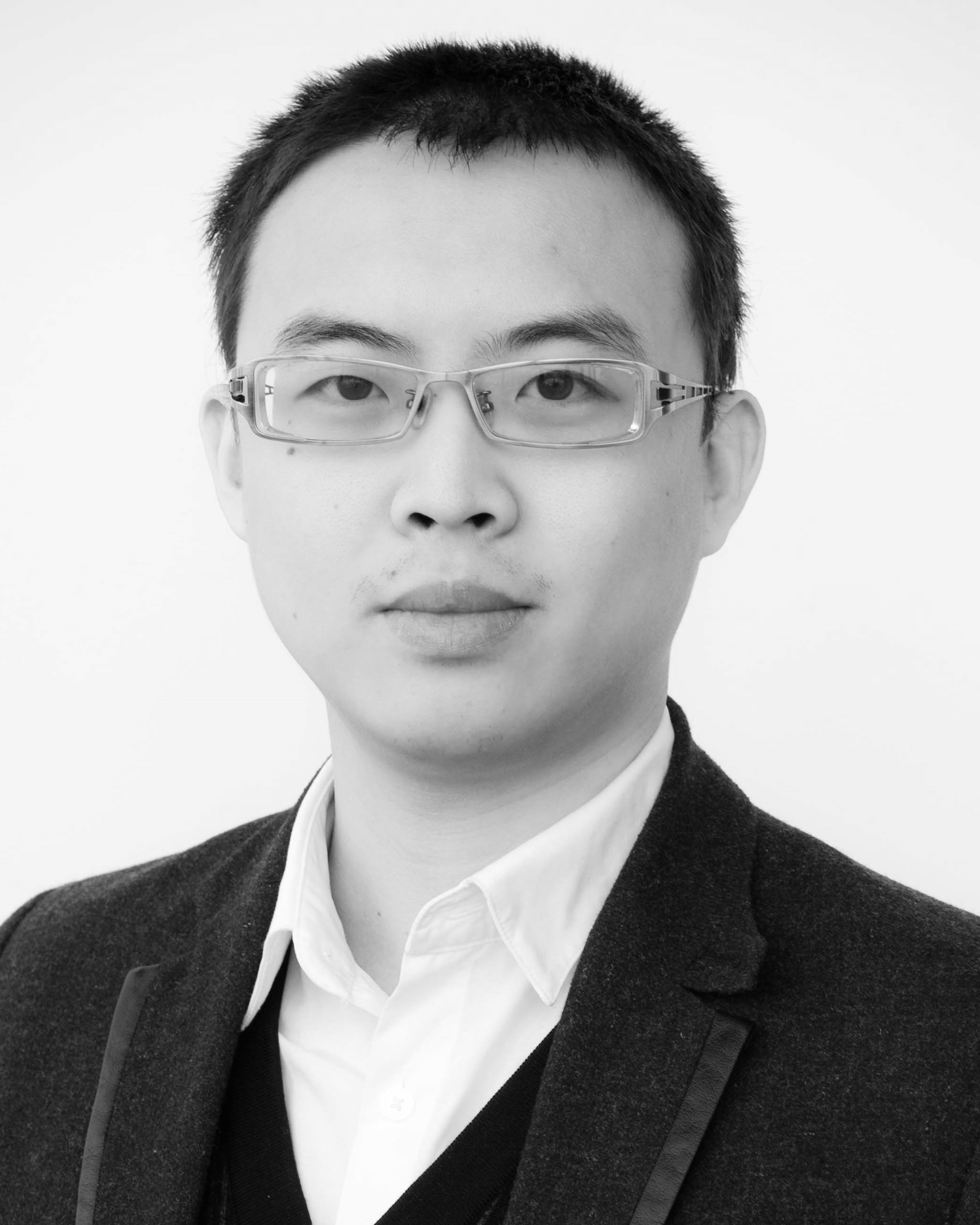}
\end{wrapfigure}
\begin{IEEEbiographynophoto}{Qi Wu}
is a postdoctoral researcher at the Australian Centre for Visual Technologies (ACVT) of the University of Adelaide. His research interests include cross-depiction object detection and classification, attributes learning, neural networks, and image captioning. He received a Bachelor in mathematical sciences from China Jiliang University, a Masters in Computer Science, and a PhD in computer vision from the University of Bath (UK) in 2012 and 2015, respectively.
\end{IEEEbiographynophoto}

\vspace{-1cm}
\begin{wrapfigure}{l}{7em}
\vspace{-10pt}
  \includegraphics[width=7em]{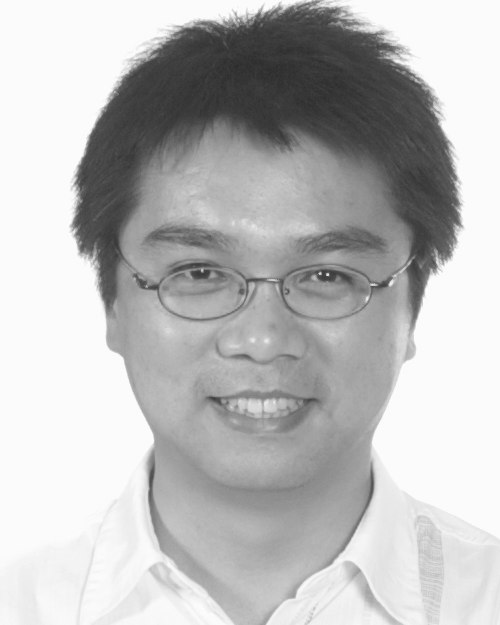}
\end{wrapfigure}
\begin{IEEEbiographynophoto}{Chunhua Shen}
is a Professor of computer science at the University of Adelaide. He was with the computer vision program at NICTA (National ICT Australia) in Canberra for six years before moving back to Adelaide. He studied at Nanjing University (China), at the Australian National University, and received his PhD degree from the University of Adelaide. In 2012, he was awarded the Australian Research Council Future Fellowship.
\end{IEEEbiographynophoto}

\vspace{-1cm}
\begin{wrapfigure}{l}{7em}
\vspace{-10pt}
  \includegraphics[width=7em]{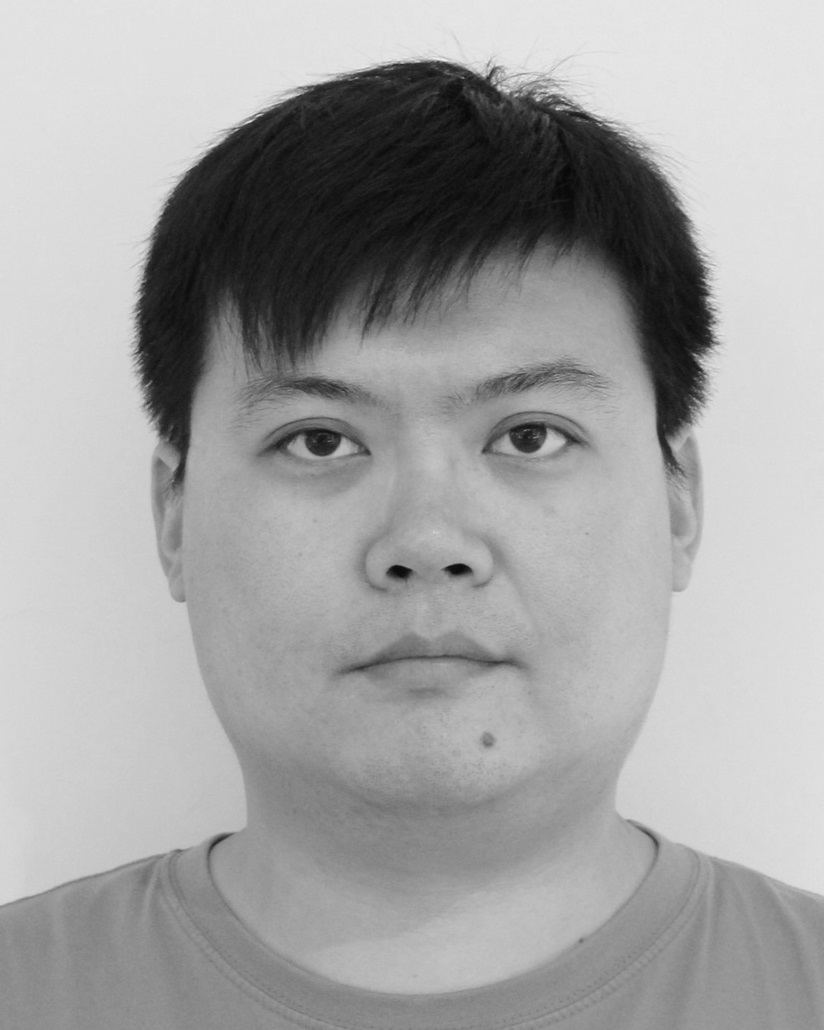}
\end{wrapfigure}
\begin{IEEEbiographynophoto}{Peng Wang}
is a postdoctoral researcher at the Australian Centre for Visual Technologies (ACVT) of the University of Adelaide. He received a Bachelor in electrical engineering and automation, and a PhD in control science and engineering from Beihang University (China) in 2004 and 2011, respectively.
\end{IEEEbiographynophoto}

\vspace{-0.5cm}
\begin{wrapfigure}{l}{7em}
\vspace{-10pt}
  \includegraphics[width=7em]{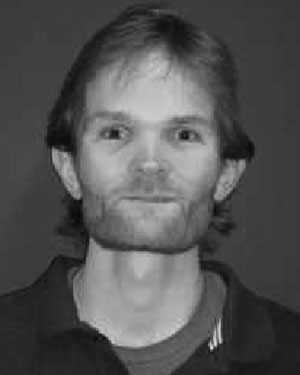}
\end{wrapfigure}
\begin{IEEEbiographynophoto}{Anthony Dick}
is an Associate Professor at the University of Adelaide. He
received a PhD degree from the University of Cambridge in 2002, where he worked on 3D reconstruction of architecture from images. His research interests include image-based modeling, automated video surveillance, and image search.
\end{IEEEbiographynophoto}

\vspace{-0.5cm}
\begin{wrapfigure}{l}{7em}
\vspace{-10pt}
  \includegraphics[width=7em]{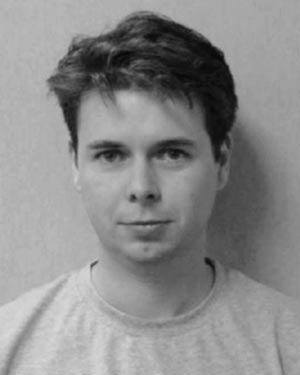}
\end{wrapfigure}
\begin{IEEEbiographynophoto}{Anton van den Hengel}
is a Professor at the University of Adelaide and the founding Director of The Australian Centre for Visual Technologies (ACVT). He received a PhD in Computer Vision in 2000, a Master Degree in Computer Science in 1994, a Bachelor of Laws in 1993, and a Bachelor of Mathematical Science in 1991, all from The University of Adelaide.
\end{IEEEbiographynophoto}

\end{document}